\documentclass{article} % For LaTeX2e
\usepackage[preprint]{colm2025_conference}

\usepackage{microtype}
\usepackage{hyperref}
\usepackage{url}
\usepackage{booktabs}
\usepackage{amsmath}
\usepackage{lineno}

\usepackage{booktabs}
\usepackage{comment}
\usepackage{graphicx}
\usepackage{subfig}
\usepackage{fancyhdr}
\usepackage{amsmath}
\usepackage{amssymb}
\usepackage{xurl}
\usepackage{adjustbox}
\usepackage{arydshln}
\usepackage{relsize}
\usepackage{wasysym}
\usepackage{multirow}
\usepackage[makeroom]{cancel}
\usepackage{scalerel,graphicx,xparse}
\usepackage{hyperref}
\usepackage{xcolor}
\usepackage{makecell}
\usepackage{tikz}
\usepackage{pgfplots}
\usepgfplotslibrary{groupplots}
\pgfplotsset{compat=1.15}
\usepackage{amsthm}
\usepackage{soul}
\usepackage[capitalize,nameinlink]{cleveref}

\newtheorem{assumption}{Assumption}

\definecolor{darkblue}{rgb}{0, 0, 0.5}
\hypersetup{colorlinks=true, citecolor=darkblue, linkcolor=darkblue, urlcolor=darkblue}

\title{On Estimating True Online Influence: Causal Modeling for Policy Evaluation and Counterfactual Prediction in Social Media}
\title{Estimating Online Influence Needs Causal Modeling!\\ Counterfactual Analysis of Social Media Engagement}

\author{Lin Tian \& Marian-Andrei Rizoiu\\
University of Technology Sydney\\
\texttt{\{Lin.Tian-3,Marian-Andrei.Rizoiu\}@uts.edu.au} \\
}

\begin{document}

\ifcolmsubmission
\linenumbers
\fi

\maketitle

\begin{abstract}
Understanding true influence in social media requires distinguishing correlation from causation—particularly when analyzing misinformation spread. While existing approaches focus on exposure metrics and network structures, they often fail to capture the causal mechanisms by which external temporal signals trigger engagement. We introduce a novel joint treatment-outcome framework that leverages existing sequential models to simultaneously adapt to both policy timing and engagement effects. Our approach adapts causal inference techniques from healthcare to estimate Average Treatment Effects (ATE) within the sequential nature of social media interactions, tackling challenges from external confounding signals. Through our experiments on real-world misinformation and disinformation datasets, we show that our models outperform existing benchmarks by 15-22\% in predicting engagement across diverse counterfactual scenarios, including exposure adjustment, timing shifts, and varied intervention durations. 
Case studies on $492$ social media users show our causal effect measure aligns strongly with the gold standard in influence estimation, the expert-based empirical influence.
\end{abstract}

% !TeX root = ../main.tex

\section{Introduction}
% In the dynamic domain of social media, a fundamental challenge lies in distinguishing mere correlation from true causal influence. 
In March 2020, as COVID-19 began its global spread, a simple graphic titled ``Flatten the Curve'' created by Associate Professor Siouxsie Wiles and cartoonist Toby Morris ignited an unprecedented cascade of social media engagement\footnote{\url{https://www.washingtonpost.com/graphics/2020/world/corona-simulator/}}. Within just 72 hours, the visualization accumulated over 10 million impressions on X (formerly Twitter), was translated into more than 25 languages, and was adopted by government health agencies worldwide \citep{bavel2020using}. What made this particular content achieve such extraordinary reach was not merely its clear visualization of pandemic dynamics---it was the complex interplay between its timely release amid escalating global concern, amplification by influential public health accounts, and concurrent spikes in search traffic as reflected in Google Trends data \citep{jackson2020hashtagactivism}. As the graphic's engagement metrics surged, a corresponding peak in ``social distancing'' searches appeared across global search indices, preceding measurable changes in mobility patterns across affected regions~\citep{gao2020association}.
Today, with hindsight, we say that the ``Flatten the Curve'' graphic was influential as it increased the public's awareness of non-pharmaceutical interventions for the COVID-19 pandemic.

But what is influence? It is more than exposure or virality; it is one's capacity to shape attitudes and change behaviors.
And while exposure (particularly repeated exposure) and influence are intimately linked, true influence estimation require causal reasoning.
% True influence in online networks requires more than mere exposure---it demands the capacity to causally shape attitudes and behaviors. 
This causal perspective on influence has been established in seminal works across multiple disciplines~\citep{aral2009distinguishing, eckles2016estimating, watts2021measuring}, which collectively demonstrate that traditional influence measures often conflate homophily with causal effects.
% To go beyond exposure and quantify proper influence requires causal frameworks rather than correlational analysis.
In online social media, information cascades equate to exposure---they are the digital equivalent of the \emph{word-of-mouth} phenomenon that allow content to spread widely.
But how do they relate to influence?
This paper introduces a causal framework to estimate user influence by conducting counterfactual analysis on misinformation and disinformation cascades.
This would allow us to understand which users, public groups or pages are truly shaping public discourse on highly polarizing topics, and who does truly drive misinformation spread.

True influence is unobserved, and notoriously difficult to estimate~\citep{ram2024empirically}.
% proper identification requires accounting for understanding information cascades necessitates causal frameworks rather than correlational analysis. 
% This distinction becomes particularly crucial when studying misinformation and disinformation, where engagement patterns can be artificially manipulated or organically amplified through complex network effects \citep{vosoughi2018spread, guess2020exposure}. 
% It underscores a critical challenge in understanding online information spread: how do external attention signals---
We therefore train and test our models on observed exogenous attention signals---like search trends, news coverage cycles, or influencer amplification---and ask how do they causally impact content engagement and information diffusion.
While existing engagement prediction approaches have made significant progress using content features and network structure~\citep{zhou2021survey}, they typically overlook the causal relationship between external temporal signals and engagement dynamics. Moreover, previous studies on influence measurement in social media have often focused on exposure rates and network structures but have not fully incorporated a causal perspective \citep{cha2010measuring,bakshy2011everyone,kwak2010twitter}. Similarly, causal inference frameworks in machine learning~\citep{pearl2009causality, peters2017elements} generally lack the temporal sophistication needed to model the intricate dependencies between time-varying external signals and sequential engagement outcomes.

To tackle these challenges, we introduce a causal framework that jointly models treatment intensities -- external signals driving engagement -- and social engagement outcomes. Our approach builds on advances in sequential modeling, adapting transformer architectures \citep{Vaswani+2017} and selective state space models (e.g.,\ Mamba \citep{gu2024mamba}) to meet the demands of causal inference with time-varying treatments. By comparing integration mechanisms -- token-based, attention-based, layer-based, and adapter-based -- we present detailed insights into the architectural requirements for effective causal modeling in social media contexts, particularly in scenarios involving misinformation.

This work explores three key research questions: (1) Does joint modeling of treatment intensities and outcomes improve predictive accuracy under realistic policy-driven scenarios, especially those aimed at curbing misinformation? (2) How effectively do transformers and state space models capture the intricate temporal dependencies between external signals and engagement outcomes? (3) How reliably can these models predict engagement outcomes under hypothetical scenarios, such as changes in exposure rates or policy active timings?

\section{Problem Statement}
\label{subsec:problem}
Let $\mathcal{E}$ denote a social media event with associated posts $\mathcal{P} = \{p_1, p_2, \ldots, p_N\}$. For each post $p \in \mathcal{P}$, we define a tuple $(t_0, x, u, o, H)$ where $t_0$ is the original posting time, $x$ represents the textual content, $u$ captures user metadata, $o \in \mathcal{O}$ indicates the content category from the set of possible categories $\mathcal{O}$, and $H = \{(t_j, e_j)\}_{j=1}^{m}$ is the interval-censored engagement history with $m$ observation intervals. Each $e_j$ is a $d$-dimensional vector capturing different types of engagement (likes, shares, comments, etc.) at observation time $t_j$.

The Google Trends information is a set of tuples $G = \{(t_k, g_k)\}_{k=1}^{l}$, with $g_k$ as a scalar representing the search intensity of keywords related to the post’s content (e.g.,\ a spike in searches for ``election'' for a political post) at time $t_k$. These signals, collected over $l$ time points, serve as an exogenous signal, capturing how real-world interest.

Given an observation window $\tau_{\text{obs}}$ starting at $t_0$, let $H_{\tau_{\text{obs}}}(p) = \{(t, e) \in H \mid t_0 \leq t \leq t_0 + \tau_{\text{obs}}\}$ represent the post’s initial engagement history, and $G_{\tau_{\text{obs}}} = \{(t, g) \in G \mid t_0 \leq t \leq t_0 + \tau_{\text{obs}}\}$ as the corresponding external signals. The goal is to predict the future engagement trajectory under policy-driven modifications to the external signal patterns (e.g.,\ adjusting Google Trends intensity to reflect shifts in public attention). Specifically, we aim to forecast $\{\hat{e}(t_0 + \tau_{\text{obs}} + k\Delta t)\}_{k=1}^{K}$, where $\Delta t$ is a fixed time step (e.g.,\ one day), and $K = \lfloor T/\Delta t \rfloor$ is the number of prediction points over a horizon $T$ (e.g.,\ one month).

\begin{figure}[h]
\centering
\begin{tikzpicture}

% Define colors
%\definecolor{lightgray}{RGB}{200,200,200}
%\definecolor{cyan}{RGB}{0,255,255}
%\definecolor{magenta}{RGB}{255,0,255}

%% (a) Observational Data - Group A (days 0 to 7)
\begin{axis}[
    name=ax1,
    width=5cm, height=3cm,
    at={(-6.7cm, 5.0cm)},  
    xlabel={Time, $t$ (days)},
    ylabel={\# Engagement},
    ylabel style={yshift=-3pt},
    ymin=0, ymax=50,
    xmin=0, xmax=7,
    xtick={0,1,2,3,4,5,6,7},
    ytick={0,10,20,30,40,50},
    grid=major,
    grid style={lightgray},
    legend style={at={(0.5,-0.7)}, anchor=north, legend columns=1}
]
\addplot[black, mark=x] coordinates {
    (0,0) (3,8) (3.5,17) (5,22) (5.5,28)
};
\addlegendentry{Engagement $o_{obs}$}
\addplot[cyan, mark=*, mark size=3] coordinates {
    (1,0) (3,0) (3.5,0) (5,0) (5.5,0) (5.75,0)
};
\addlegendentry{Events $a_{obs}$}
\end{axis}

\begin{axis}[
    name=ax2,
    at={(-6.7cm, 1.2cm)},  
    width=5cm, height=3cm,
    xlabel={Time, $t$ (days)},
    ylabel={\# Engagement},
    ymin=0, ymax=10,
    xmin=0, xmax=7,
    xtick={0,1,2,3,4,5,6,7},
    ytick={0,2,4,6,8,10},
    grid=major,
    grid style={lightgray},
    legend style={at={(0.5,-0.7)}, anchor=north, legend columns=2}
]
\addplot[blue, thick, mark=x] coordinates {
    (0,0) (3,4) (3.5,5) (5,7) (5.5,8)
};
\addlegendentry{Likes}
\addplot[red, thick, mark=square] coordinates {
    (0,0) (3,2) (3.5,5) (5,6) (5.5,9)
};
\addlegendentry{Shares}
\addplot[brown, thick, mark=triangle] coordinates {
    (0,0) (3,0) (3.5,4) (5,4) (5.5,6)
};
\addlegendentry{Comments}
\addplot[purple, thick, mark=diamond] coordinates {
    (0,0) (3,2) (3.5,3) (5,5) (5.5,5)
};
\addlegendentry{Emojis}

\end{axis}

% (b) Interventional Query - Cumulative Engagement (days 0 to 14)
\begin{axis}[
    name=ax3,
    width=5cm, height=3cm,
    at={(-2.1cm, 5.0cm)},  
    xlabel={Time, $t$ (days)},
    ylabel={Engagement $f$},
    ylabel style={yshift=-7pt},
    ymin=0, ymax=150,
    xmin=0, xmax=14,
    xtick={0,2,4,6,7,8,10,12,14},
    ytick={0,30,60,90,120,150},
    grid=major,
    grid style={lightgray},
    legend style={at={(0.5,-0.7)}, anchor=north, legend columns=2}
]
% Observed engagement: solid blue line
\addplot[blue, thick] coordinates {
    (0,0) (3,8) (3.5,17) (5,22) (5.5,28) (6,28) (7,28)
};
\addlegendentry{$f_{obs}$}

\addplot[blue, thick, dashed] coordinates {
    (7,28) (8,32.2) (9,36.4) (10,40.6) (11,44.8) (12,49.0) (13,53.2) (14,57.4)
};
\addlegendentry{$f_{\text{pred}}$}

% Predicted engagement under pi_A: dashed red line
\addplot[red, thick] coordinates {
    (7,28) (8,39.2) (9,44.8) (10,44.8) (11,57.4) (12,70.0) (13,77.0) (14,95.2)
};
\addlegendentry{$f_{\pi_A}$}
% Add vertical line at day 7 to mark intervention start
\draw[gray, thin] (7,0) -- (7,10);
\end{axis}

% (b) Interventional Query - Normalized Intensity (days 0 to 14)
\begin{axis}[
    name=ax4,
    width=5cm, height=3cm,
    at={(-2.1cm, 1.2cm)},  
    xlabel={Time, $t$ (days)},
    ylabel={Intensity $\lambda$},
    ylabel style={yshift=-7pt},
    ymin=0, ymax=1,
    xmin=0, xmax=14,
    xtick={0,2,4,6,7,8,10,12,14},
    ytick={0,0.2,0.4,0.6,0.8,1.0},
    grid=major,
    grid style={lightgray},
    legend style={at={(0.5,-0.7)}, anchor=north, legend columns=2}
]

\addplot[green,  line width=1.5pt, dashdotted] coordinates {
    (0,0.2) (1,0.3) (2,0.25) (3,0.35) (4,0.4) (5,0.3) (6,0.35) (7,0.2)
};
\addlegendentry{$\lambda_{obs}$}
\addplot[red,  line width=1.5pt, dashed] coordinates {
    (7,0.2) (8,1.0) (9,1.0) (10,1.0) (11,0.8) (12,0.6) (13,0.5) (14,0.5)};
\addlegendentry{$[\pi_A]$}
\addplot[cyan, mark=*, mark size=3] coordinates {
    (1,0) (3,0) (3.5,0) (5,0) (5.5,0) (5.75,0)
};
\addplot[magenta, mark=*, mark size=3] coordinates {
    (8,0) (9,0) (10,0) (11,0) (12,0) (13,0) (14,0)
};
\addplot[magenta, mark=*, mark size=4.5] coordinates {
    (8,0) (9,0)
};
\addplot[fill=lightgray, opacity=0.3] coordinates {
    (7,0) (8,0) (8,1) (7,1) (7,0)
};
% Add vertical line at day 7 to mark intervention start
\draw[gray, thin] (7,0) -- (7,10);
\end{axis}

% (c) Counterfactual Query - Cumulative Engagement (days 0 to 10)
\begin{axis}[
    name=ax5,
    width=5cm, height=3cm,
    at={(2.6cm, 5cm)},  
    xlabel={Time, $t$ (days)},
    ylabel={Engagement $f$},
    ylabel style={yshift=-7pt},
    ymin=0, ymax=150,
    xmin=0, xmax=14,
    xtick={0,2,4,5,6,7,8,10,12,14},
    ytick={0,30,60,90,120,150},
    grid=major,
    grid style={lightgray},
    legend style={at={(0.5,-0.7)}, anchor=north, legend columns=2}
]
% Observed engagement: solid blue line
\addplot[blue, thick] coordinates {
    (0,0) (3,8) (3.5,17) (5,22) (5.5,28) (6,28) (7,28)
};
\addlegendentry{$f_{obs}$}

\addplot[blue, thick, dashed] coordinates {
    (7,28) (8,32.2) (9,36.4) (10,40.6) (11,44.8) (12,49.0) (13,53.2) (14,57.4)
};
\addlegendentry{$f_{\text{pred}}$}

% Predicted engagement under pi_A: dashed red line
\addplot[red, thick] coordinates {
    (7,28) (8,39.2) (9,44.8) (10,44.8) (11,57.4) (12,70.0) (13,77.0) (14,95.2)
};
\addlegendentry{$f_{\pi_A}$}
% Another line for f_int under pi_B (alternative prediction): solid red line
\addplot[orange, thick] coordinates {
    (7,28) (8,57.6) (9,72.8) (10,80.8) (11,99.6) (12,101.8) (13,116.0) (14,129.2)
};
\addlegendentry{$f_{\pi_B}$}
% Add vertical line at day 7 to mark intervention start
\draw[gray, thin] (7,0) -- (7,10);
\end{axis}

% (c) Counterfactual Signal - Normalized Intensity (days 0 to 10)
\begin{axis}[
    name=ax6,
    width=5cm, height=3cm,
    at={(2.6cm, 1.2cm)},  
    xlabel={Time, $t$ (days)},
    ylabel={Intensity $\lambda$},
    ylabel style={yshift=-7pt},
    ymin=0, ymax=1,
    xmin=0, xmax=14,
    xtick={0,2,4,5,6,7,8,10,12,14},
    ytick={0,0.2,0.4,0.6,0.8,1.0},
    grid=major,
    grid style={lightgray},
    legend style={at={(0.5,-0.7)}, anchor=north, legend columns=2}
]

\addplot[green,  line width=1.5pt, dashdotted] coordinates {
    (0,0.2) (1,0.3) (2,0.25) (3,0.35) (4,0.4) (5,0.3) (6,0.35) (7,0.2)
};
\addlegendentry{$\lambda_{obs}$}
\addplot[orange,  line width=1.5pt, dashed] coordinates {
    (7,0.2) (8,1.0) (9,1.0) (10,1.0) (11,0.8) (12,0.6) (13,0.5) (14,0.5)
};
\addlegendentry{$[\pi_B]$}
\addplot[cyan, mark=*, mark size=3] coordinates {
    (1,0) (3,0) (3.5,0) (5,0) (5.5,0) (5.75,0)
};
\addplot[magenta, mark=*, mark size=3] coordinates {
    (8,0) (9,0) (10,0) (11,0) (12,0) (13,0) (14,0)
};
\addplot[magenta, mark=*, mark size=4.5] coordinates {
    (8,0) (9,0)
};
\addplot[fill=lightgray, opacity=0.3] coordinates {
    (7,0) (10,0) (10,1) (7,1) (7,0)
};
% Add vertical line at day 7 to mark intervention start
\draw[gray, thin] (7,0) -- (7,10);
\end{axis}

\node at (-5cm, 7cm) {(a) Observational data};  
\node at (-0.5cm, 7cm) {(b) Exogenous signals};  
\node at (4.0cm, 7cm) {(c) Counterfactual signals};  

\end{tikzpicture}
\caption{Visualization of engagement data and queries for social media post $p$.
\textbf{(a) Observational data during the period $[0,7]$ days.} The top plot shows cumulative engagement over time (black line with crosses) and observed events (cyan dots). The bottom plot displays individual engagement metrics: Likes (blue crosses), Shares (red squares), Comments (brown triangles), and Emojis (purple diamonds).
\textbf{(b) The exogenous signal.} The engagement trajectory of post $p$ after the observation period under policy $\pi_A$, derived from Google Trends data, with a one-day exposure time (shaded area from day 7 to 8). The top plot shows observed engagement (solid blue line), predicted engagement without intervention (dashed blue line), and predicted engagement under $\pi_A$ (dashed red line). The bottom plot displays normalized intensity $\lambda_{obs}$ (dashed green line), policy $\pi_A$ intensity (dashed red line), observed events (cyan dots), and policy actions (magenta dots). A vertical line at day 7 marks the intervention start. 
\textbf{(c) The counterfactual signal.} How the engagement trajectory of post $p$ would have evolved if policy $\pi_B$ had been applied during $[0,7]$ with a three-day exposure time (shaded area from day 7 to 10). The top plot shows observed engagement (solid blue line), predicted engagement without intervention (dashed blue line), predicted engagement under $\pi_A$ (dashed red line), and counterfactual engagement under $\pi_B$ (solid orange line). The bottom plot displays normalized intensity $\lambda_{obs}$(dashed green line), counterfactual policy $\pi_B$ intensity (dashed orange line), observed events (cyan dots), and policy actions (magenta dots). A vertical line at day 7 marks the intervention start.
}
\label{fig:framework}
\end{figure}
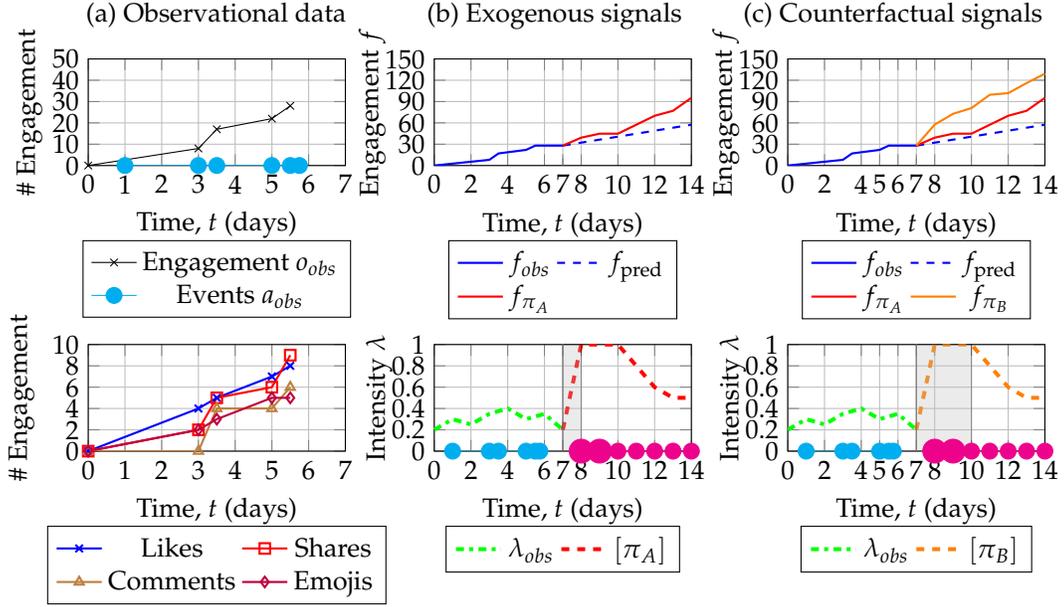
\section{Methodology}
\label{sec:methodology}
Our framework (as shown in \cref{fig:framework}) models external signals as continuous-intensity treatments through modeling that capture both immediate responses and long-range dependencies. 
We achieve this by jointly modeling the bidirectional relationship between external signals and engagement outcomes.

\subsection{Joint Treatment-Outcome Model}
\label{subsec:joint_model}
Inspired by the causal modeling framework of \citet{hizli2023causal}, we propose a joint treatment-outcome model tailored for predicting social engagement in discrete-time social media data. While \citet{hizli2023causal} use marked point processes for treatment intensity and conditional Gaussian processes for continuous-time outcomes, we adapt this approach to discrete time using deep sequential models. This adaptation improves scalability and flexibility for large-scale, irregularly sampled social media datasets while preserving the bidirectional dependency between treatments (external signals) and engagement outcomes.

\textbf{Treatment Intensity Modeling}
\label{subsubsec:treatment_model}
We model treatment intensity $\lambda^*_\pi(t)$ as the probability of a binary treatment event (e.g., a Google Trends spike) at each discrete time step $t$. The intensity is computed via a square transformation of a latent function $g^*_\pi(t)$:
\begin{equation*}
\lambda^*_\pi(t) = \left(\beta_0 + g_b(t) + g^*_a(t) + g^*_o(t) + g^*_g(t)\right)^2,
\end{equation*}
where $\beta_0$ represents a constant baseline intensity, $g_b(t)$ is a time-dependent baseline function, $g^*_a(t)$ models dependence on past treatments, $g^*_o(t)$ captures dependence on past engagement outcomes, and $g^*_g(t)$ incorporates Google Trends intensity.

For $g^*_g(t)$, we use discrete-time window sampling at 10-minute intervals ($\Delta_g = 10$ minutes):
\begin{equation*}
g^*_g(t) = \sum_{k=0}^{w-1} \alpha_k \cdot g(t - k\Delta_g) \cdot \mathbf{1}_{[t - k\Delta_g, t - (k-1)\Delta_g)}(t),
\end{equation*}
Here, $w$ is the number of historical windows considered, $g(t - k\Delta_g)$ is the Google Trends value at time $t - k\Delta_g$, $\alpha_k$ are learnable decay coefficients, and $\mathbf{1}_{[a,b)}(t)$ is an indicator function. $\lambda^*_\pi(t)$ is evaluated at discrete time steps, with treatments sampled as binary events based on this probability.

To align Google Trends with engagement observations, we implement a temporal alignment mechanism. For each observation time $t_j$, we collect signals within a lag window:
\begin{equation*}
G(t_j) = \{g_k \mid t_j - \tau_{\text{lag}} \leq t_k < t_j\}
\end{equation*}
where $\tau_{\text{lag}}$ is a hyperparameter determined through cross-validation. 
The resulting feature vector is $\mathbf{v}_g(t_j) = [g(t_j - \Delta_g), g(t_j - 2\Delta_g), \ldots, g(t_j - w\Delta_g)]$.

\subsection{Causal Assumptions}
\label{subsec:causal_assumptions}
Our causal framework relies three key assumptions suitable for social media contexts:
\begin{assumption}[Consistency] 
The potential engagement outcome $Y[a]$ under a specific pattern of external signals $a$ equals the observed engagement $Y$ when those exact signals actually occur.
\end{assumption}
% This assumption simply states that if we observe certain Google Trends patterns and corresponding engagement, this same relationship would hold if those patterns reoccur. 
This assumes observed Google Trends-engagement relationships persist if patterns reoccur.
In social media environments, this consistency is supported by the relative stability of engagement mechanisms---including recommendation algorithms, user interfaces, and notification systems---which function as reliable mediators between external signals and user behavior~\citep{becker2017network}. 
Empirical evidence from \citet{calderon2024opinion} reinforces this view, as their Opinion Market Model (OMM) incorporating Google Trends signals consistently outperformed baselines in predictive accuracy by approximately 17\% lower error rates. 
Similarly, \citet{rizoiu2017expecting} demonstrated superior popularity prediction results with their HIP model by assuming that viewcount is externally driven by sharing behavior~\citep{muchnik2013social} and explicitly accounting for that as an exogenous driver.
% further validating the stability of these signal-engagement relationships. 
They prove that HIP has the Linear Time-Invariant property which ensures that identical external stimuli produce consistent engagement responses regardless of when they occur.
While acknowledging temporal constraints, we assert this assumption holds within reasonable timeframes where platform mechanics and user behavior patterns remain stable, prior to significant algorithm updates or shifts in engagement norms.

\begin{assumption}[Fully-Mediated Policy Effect]
External signals (like Google Trends spikes) affect engagement outcomes through observable mechanisms rather than hidden pathways.
\end{assumption}
This assumption states that external signals (like Google Trends spikes) affect engagement outcomes through observable mechanisms rather than hidden pathways~\citep{bakshy2015exposure}.
The causal process operates through a well-defined mediational sequence: 
exogenous search intensity signals $\rightarrow$ heightened topical salience $\rightarrow$ content exposure via platform mechanisms $\rightarrow$ measurable engagement behaviors (e.g.,\ likes, shares, comments).
% This pathway is empirically validated by \citet{calderon2024opinion}, whose two-tier Opinion Market Model demonstrated how external signals influence both the size of the attention market and its distribution among competing opinions. 
\citet{calderon2024opinion} empirically validated this pathway, showing that external signals affect both the attention market size and opinion distribution.
Similarly, \citet{bakshy2015exposure} provide further evidence by showing how exposure mechanisms directly mediate the relationship between content availability and user engagement with diverse information.
\citet{rizoiu2017expecting} quantifies this relationship through their ``popularity impulse response'' function, measuring how external signals directly translate to viewership through measurable platform dynamics.
Their simulated intervention experiments show that media coverage predictably shapes opinion market shares through these mechanisms.
While alternative pathways might exist--such as algorithmic amplification or coordinated network activity--these appear to be minimal, as evidenced by both OMM's and HIP's superior predictive performance when directly modeling this attention-mediated relationship.

\begin{assumption}[Temporal Precedence]
Causes precede effects - current engagement can be influenced by past external signals but not by future ones.
\end{assumption}
This assumption aligns with the natural temporal ordering of social media interactions, where content engagement follows rather than precedes external attention signals. 
\citet{Rizoiu2017b} provide strong empirical support through their temporally-sensitive Hawkes Intensity Process model, which demonstrates that popularity dynamics strictly follow the causal direction where external stimuli (like promotions) at time $t$ can only influence engagement at times $\tau > t$, with a clear measurable \emph{maturity time} quantifying this temporal lag.
This maturity time metric captures the delay between external stimuli and the resulting engagement effects, providing a quantitative measure of the temporal precedence relationship.
Their model's success in predicting YouTube viewcounts using this temporal framework validates the directional nature of the causal relationship. 
Similarly, \citet{calderon2024opinion} validate this in their Opinion Market Model, showing improved predictive performance when treating search trends as antecedent to engagement spikes. 
This temporal structure is further reinforced by platform mechanics, as content discovery algorithms typically respond to rising search interest by increasing content visibility, creating a causal chain that is unidirectional in time~\citep{centola2018experimental}.

These assumptions, while not exhaustively verifiable, provide a reasonable foundation for causal inference in social media settings and allow us to identify treatment effects from observational data.
Thus, building on the potential outcomes framework, we formulate our causal inference approach to estimate the effects of interventions on engagement trajectories.

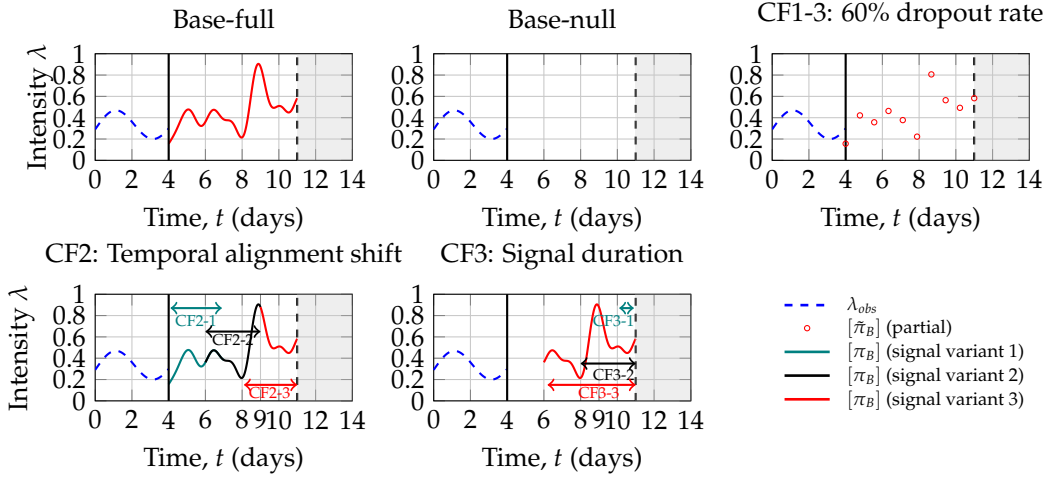
\begin{figure}
\centering
\begin{tikzpicture}

% Define colors
\definecolor{lightgray}{RGB}{200,200,200}
\definecolor{cyan}{RGB}{0,255,255}
\definecolor{magenta}{RGB}{255,0,255}

% Base case
\begin{axis}[
    name=base_full,
    width=5cm, height=3cm,
    at={(0cm, 3.2cm)},
    xlabel={Time, $t$ (days)},
    ylabel={Intensity $\lambda$},
    ylabel style={yshift=-7pt},
    ymin=0, ymax=1,
    xmin=0, xmax=14,
    xtick={0,2,4,6,8,10,12,14},
    ytick={0,0.2,0.4,0.6,0.8,1.0},
    grid=major,
    grid style={lightgray},
    title={Base-full}
]
% Historical data - engagement curve (blue)
\addplot[blue, thick, dashed, domain=0:4, samples=50] {0.25 + 0.15*sin(deg(1.5*x)) + 0.1*exp(-((x-3)^2)/10)};

% Google Trends signals after the line (red) - continue from the last blue point
\addplot[red, thick, domain=4:11, samples=300] {
    0.15 + 0.15*sin(deg(1.5*x)) + 0.1*exp(-((x-8)^2)/10) + 
    (0.3 + 0.2*sin(deg(3*x)) + 0.1*cos(deg(5*x))) * (1 / (1 + exp(-(x - 6))))
};

% Vertical line at start of prediction period
\draw[thick] (4,0) -- (4,1);

% Prediction vertical line (dashed)
\draw[thick, dashed] (11,0) -- (11,1);

% Gray background for prediction region
\fill[lightgray, opacity=0.3] (11,0) rectangle (14,1);

% Add 7 days label
% \draw[gray, thin] (6,0) -- (6,1);
% \draw[gray, thin] (7,0) -- (7,1);
% \node at (6.5,-0.15) {7 days};

\end{axis}

% Base case
\begin{axis}[
    name=base,
    width=5cm, height=3cm,
    at={(4.5cm, 3.2cm)},
    xlabel={Time, $t$ (days)},
    ylabel={},
    ylabel style={yshift=-7pt},
    ymin=0, ymax=1,
    xmin=0, xmax=14,
    xtick={0,2,4,6,8,10,12,14},
    ytick={0,0.2,0.4,0.6,0.8,1.0},
    grid=major,
    grid style={lightgray},
    title={Base-null}
]
% Historical data - engagement curve (blue)
\addplot[blue, thick, dashed, domain=0:4, samples=50] {0.25 + 0.15*sin(deg(1.5*x)) + 0.1*exp(-((x-3)^2)/10)};

% Vertical line at start of prediction period
\draw[thick] (4,0) -- (4,1);

% Prediction vertical line (dashed)
\draw[thick, dashed] (11,0) -- (11,1);

% Gray background for prediction region
\fill[lightgray, opacity=0.3] (11,0) rectangle (14,1);

% % Add 7 days label
% \draw[gray, thin] (6,0) -- (6,1);
% \draw[gray, thin] (7,0) -- (7,1);
% \node at (6.5,-0.15) {7 days};

\end{axis}

% CF1 - with 60% fluctuation
\begin{axis}[
    name=cf1,
    width=5cm, height=3cm,
    at={(9cm, 3.2cm)},
    xlabel={Time, $t$ (days)},
    ylabel={},
    ymin=0, ymax=1,
    xmin=0, xmax=14,
    xtick={0,2,4,6,8,10,12,14},
    ytick={0,0.2,0.4,0.6,0.8,1.0},
    grid=major,
    grid style={lightgray},
    title={CF1-3: 60\% dropout rate}
]
% Historical data curve (blue)
\addplot[blue, thick, dashed, domain=0:4, samples=50] {0.25 + 0.15*sin(deg(1.5*x)) + 0.1*exp(-((x-3)^2)/10)};

\addplot[
    red,
    only marks,
    mark=o,           % open circle
    mark size=1pt,    % even smaller
    samples=10,
    domain=4:11
] {
    0.15 + 0.15*sin(deg(1.5*x)) +
    0.1*exp(-((x-8)^2)/10) +
    (0.3 + 0.2*sin(deg(3*x)) + 0.1*cos(deg(5*x))) * (1 / (1 + exp(-(x - 6))))
};

% Vertical line at start of prediction period
\draw[thick] (4,0) -- (4,1);

% Prediction vertical line (dashed)
\draw[thick, dashed] (11,0) -- (11,1);

% Gray background for prediction region
\fill[lightgray, opacity=0.3] (11,0) rectangle (14,1);

% Add 7 days label
% \draw[gray, thin] (6,0) -- (6,1);
% \draw[gray, thin] (7,0) -- (7,1);
% \node at (6.5,-0.15) {7 days};

\end{axis}

% CF2 - with 2d highlight
\begin{axis}[
    name=cf2,
    width=5cm, height=3cm,
    at={(0cm, 0cm)},
    xlabel={Time, $t$ (days)},
    ylabel={Intensity $\lambda$},
    ymin=0, ymax=1,
    xmin=0, xmax=14,
    xtick={0,2,4,6,8,9,10,12,14},
    ytick={0,0.2,0.4,0.6,0.8,1.0},
    grid=major,
    grid style={lightgray},
    title={CF2: Temporal alignment shift}
]
% Historical data with some fluctuations (blue)
\addplot[blue, thick, dashed, domain=0:4, samples=50] {0.25 + 0.15*sin(deg(1.5*x)) + 0.1*exp(-((x-3)^2)/10)};

\addplot[teal, thick, domain=4:7, samples=300] {
    0.15 + 0.15*sin(deg(1.5*x)) +
    0.1*exp(-((x-8)^2)/10) +
    (0.3 + 0.2*sin(deg(3*x)) + 0.1*cos(deg(5*x))) * (1 / (1 + exp(-(x - 6))))
};

\addplot[black, thick, domain=6:9, samples=300] {
    0.15 + 0.15*sin(deg(1.5*x)) +
    0.1*exp(-((x-8)^2)/10) +
    (0.3 + 0.2*sin(deg(3*x)) + 0.1*cos(deg(5*x))) * (1 / (1 + exp(-(x - 6))))
};

\addplot[red, thick, domain=9:11, samples=300] {
    0.15 + 0.15*sin(deg(1.5*x)) +
    0.1*exp(-((x-8)^2)/10) +
    (0.3 + 0.2*sin(deg(3*x)) + 0.1*cos(deg(5*x))) * (1 / (1 + exp(-(x - 6))))
};

% Vertical line at start of prediction period
\draw[thick] (4,0) -- (4,1);

% 2d region with highlighted area
% \fill[lightgray, opacity=0.3] (7.5,0) rectangle (8,1);

% Vertical line at start of 2d period
% \draw[thick, dashed] (7.5,0) -- (7.5,1);

% Prediction vertical line
\draw[thick, dashed] (11,0) -- (11,1);

% Gray background for prediction region
\fill[lightgray, opacity=0.3] (11,0) rectangle (14,1);
% === Annotations for CF2-1 ===
\draw[<->, thick, teal] (axis cs:4.1,0.87) -- (axis cs:6.9,0.87);
\node[teal] at (axis cs:5.5,0.75) {\tiny CF2-1};

% === Annotations for CF2-2 ===
\draw[<->, thick, black] (axis cs:6,0.65) -- (axis cs:9,0.65);
\node[black] at (axis cs:7.5,0.58) {\tiny CF2-2};

% === Annotations for CF2-3 ===
\draw[<->, thick, red] (axis cs:8.1,0.15) -- (axis cs:11,0.15);
\node[red] at (axis cs:9.5,0.08) {\tiny CF2-3};

\end{axis}

% CF3 - with k3d peak
\begin{axis}[
    name=cf3,
    width=5cm, height=3cm,
    at={(4.5cm, 0cm)},
    xlabel={Time, $t$ (days)},
    ylabel={},
    ylabel style={yshift=-7pt},
    ymin=0, ymax=1,
    xmin=0, xmax=14,
    xtick={0,2,4,6,8,9,10,12,14},
    ytick={0,0.2,0.4,0.6,0.8,1.0},
    grid=major,
    grid style={lightgray},
    title={CF3: Signal duration}
]
% Historical data curve (blue)
\addplot[blue, thick, dashed, domain=0:4, samples=50] {0.25 + 0.15*sin(deg(1.5*x)) + 0.1*exp(-((x-3)^2)/10)};

% Google Trends signals with peak in k3d region and noise (red) - continue from the last blue point
\addplot[red, thick, domain=6:11, samples=300] {
    0.15 + 0.15*sin(deg(1.5*x)) +
    0.1*exp(-((x-8)^2)/10) +
    (0.3 + 0.2*sin(deg(3*x)) + 0.1*cos(deg(5*x))) * (1 / (1 + exp(-(x - 6))))
};

% Vertical line at start of prediction period
\draw[thick] (4,0) -- (4,1);

% Vertical line at end of k3d period
% \draw[thick, dashed] (7,0) -- (7,1);

% Prediction vertical line
\draw[thick, dashed] (11,0) -- (11,1);

% Gray background for prediction region
\fill[lightgray, opacity=0.3] (11,0) rectangle (14,1);

% === Annotations for CF3-1 ===
\draw[<->, thick, teal] (axis cs:10.1,0.87) -- (axis cs:10.9,0.87);
\node[teal] at (axis cs:9.8,0.75) {\tiny CF3-1};

% === Annotations for CF3-2 ===
\draw[<->, thick, black] (axis cs:8,0.35) -- (axis cs:11,0.35);
\node[black] at (axis cs:9.8,0.25) {\tiny CF3-2};

% === Annotations for CF3-3 ===
\draw[<->, thick, red] (axis cs:6.2,0.15) -- (axis cs:11,0.15);
\node[red] at (axis cs:9.0,0.08) {\tiny CF3-3};

\end{axis}

\begin{axis}[
    hide axis,
    scale only axis,
    height=0cm,
    width=0cm,
    at={(12.5cm, 1.5cm)},
    anchor=north east,
    legend style={
        draw=none,
        column sep=5pt,
        font=\scriptsize,
        nodes={inner sep=1pt},
    },
    legend cell align=left,
    legend columns=1,
    legend entries={
        {$\lambda_{obs}$},
        {$[\tilde{\pi}_B]$ (partial)},
        {$[\pi_B]$ (signal variant 1)},
        {$[\pi_B]$ (signal variant 2)},
        {$[\pi_B]$ (signal variant 3)}
    },
]
\addplot[blue, dashed, line width=1pt] {0};          % lambda_obs
\addlegendimage{only marks, red, mark=o, mark size=1pt};    % partial pi_B
\addlegendimage{teal, line width=1pt};                      % solid variant 1
\addlegendimage{black, line width=1pt};                     % solid variant 2
\addlegendimage{red, line width=1pt};                       % solid variant 3
\end{axis}

\end{tikzpicture}
\caption{
Engagement trajectory and counterfactual scenarios. 
Blue dashed lines represent observed social media engagement ($\lambda_{obs}$).
Red lines indicate Google Trends signals ($[\pi_B]$).
$\tilde{\pi}_B$ as the partial observed Google Trends signals.
The vertical dashed line at day 9 marks the start of prediction period, with the gray shaded area showing the prediction region.
}
\label{fig:cf_scenarios}
\end{figure}

\subsection{Counterfactual Analysis}
\label{subsubsec:counterfactual}
We introduce a counterfactual analysis framework that systematically manipulates exogenous signals along the temporal dimension to understand their causal impact on engagement dynamics, as shown in Figure~\ref{fig:cf_scenarios}.
Formally, a counterfactual scenario $\mathcal{C}$ as a transformation of the Google Trends signal $G = \{(t_k, g_k)\}_{k=1}^{l}$ through a temporal manipulation function $\Psi_{\theta}$:
\begin{equation*}
G_{\mathcal{C}} = \Psi_{\theta}(G) = \{(t_k + \delta_{\theta}(t_k, g_k), g_k \cdot \gamma_{\theta}(t_k, g_k))\}_{k=1}^{l},
\end{equation*}
where $\delta_{\theta}$ shifts the timing of signals and $\gamma_{\theta}$ adjusts signal intensity, enabling three counterfactual scenarios (CF): Dropout Rate (CF1) adjusts intensity ($\gamma_{\theta}$) to explore exposure effects (0\%~$\rightarrow$20\% as CF1-1, 20\%$\rightarrow$40\% as CF1-2, 40\%$\rightarrow$~60\% as CF1-3); Temporal Alignment Shifts (CF2) the onset timing ($\delta_{\theta}$) of a fixed-duration signal to test sensitivity to early exposure; Signal Duration Manipulation (CF3) modifies persistence to evaluate sustained attention effects (1-day as CF3-1, 3-day as CF3-2, 5-day as CF3-3).
For each counterfactual scenario, we estimate the expected engagement outcomes under the transformed signal and calculate the causal effect as the difference between counterfactual and actual outcomes: $\Delta_{\mathcal{C}} = \mathbb{E}[Y | G_{\mathcal{C}}] - \mathbb{E}[Y | G]$.

\subsection{Model Training and Optimization}
\label{subsec:training}
We train our joint model using a combined loss function that explicitly optimizes both treatment intensity modeling and outcome prediction:

\begin{equation}
\mathcal{L}_{\text{joint}} = \underbrace{\text{MSE}(Y_{\text{true}}(t), Y_{\text{pred}}(t;\theta_Y))}_{\text{Outcome Loss}} + \alpha \underbrace{\text{BCE}(A_{\text{true}}(t), \lambda_A(t; \theta_A))}_{\text{Intensity Loss}},
\end{equation}

where $\alpha$ is a hyperparameter controlling the relative importance between accurate outcome prediction and accurate treatment intensity modeling. This joint optimization ensures that the model captures both the occurrence pattern of external signals (treatments) and their effects on engagement outcomes.

For the Mamba-based model, we additionally incorporate a temporal coherence loss, which ensures consistent state transitions across varying time intervals with
$\mathcal{L}_{\text{temp}} = \frac{1}{|\mathcal{P}|} \sum_{p \in \mathcal{P}} \sum_{j=0}^{m-1} \|\mathbf{h}_{j+1} - \exp(\Delta t^+_j \cdot \tilde{\mathbf{A}}_t)\mathbf{h}_j\|^2.
$
The temporal coherence loss regularizes the hidden state dynamics to follow the theoretical state transitions defined by the continuous-time state space model, where $\mathbf{h}_j$ is the hidden state at time $t_j$, $\Delta t^+_j = t_{j+1} - t_j$ is the forward time interval, and $\exp(\Delta t^+_j \cdot \tilde{\mathbf{A}}_t)$ represents the state transition matrix over the interval $\Delta t^+_j$. By minimizing this loss, we ensure that the model maintains consistent internal representations despite the irregular sampling intervals.

For the transformer-based model, we replace the temporal coherence loss with an attention consistency loss that encourages similar attention patterns for events with similar temporal relationships with 
$
\mathcal{L}_{\text{att}} = \frac{1}{|\mathcal{P}|} \sum_{p \in \mathcal{P}} \sum_{i,j,k,l} w_{ijkl} \|\mathbf{A}_{ij} - \mathbf{A}_{kl}\|^2.
$
where $\mathbf{A}_{ij}$ represents the attention weight from position $i$ to position $j$, and $w_{ijkl}$ is a similarity weight based on the temporal distance between positions.

% \subsection{Two-Tier Architecture for Group-Level Dynamics}
% \label{subsec:two_tier}

% To model both individual post engagement and collective category engagement, we implement a hierarchical two-tier architecture. The first tier processes individual posts using our joint treatment-outcome model, while the second tier captures inter-post temporal dependencies within the same category.

% For each category $o \in \mathcal{O}$, we process all posts $p_i \in \mathcal{P}_o$ individually using the first-tier model:

% \begin{equation}
% \mathbf{h}_i = \text{Model}_1(H_{\tau_{obs}}(p_i), G_{\tau_{obs}}, x_i, u_i), \quad \forall p_i \in \mathcal{P}_o
% \end{equation}

% Then, we model the temporal interactions between posts sharing the same category in the second tier:

% \begin{equation}
% \mathbf{z}_o = \text{Model}_2((\mathbf{h}_i, \delta t_i))
% \end{equation}

% where $\delta t_i = t^p_{i+1} - t^p_i$ represents the inter-post intervals, and posts are ordered chronologically by posting time $t^p_i$.

\section{Experiment and Results}
This section introduces the datasets, models and baselines, and report our experimental findings on the optimal architecture choice for exogenously-driven engagement modeling.
\subsection{Datasets}
\textbf{Social Media Engagement Data.}
The SocialSense dataset~\citep{kong2022slipping} is a misinformation dataset containing public Facebook posts collected between 2019 and 2021. It spans four domains: Australian Bushfires (78,030 posts), Climate Change (138,278 posts), Vaccination (178,894 posts), and COVID-19 (640,100 posts). These posts, labeled by domain experts, focus on misinformation and conspiratorial discussions. User engagements (likes, shares, comments, and emoji) were collected using the CrowdTangle API~\footnote{\url{https://www.crowdtangle.com/} before its termination in August 2024.}. 

The DiN dataset~\citep{tian2025before} is a disinformation dataset comprising $746,653$ posts from $41$ accounts from 2019 to 2024. It is identified by social science experts as part of coordinated information operations. The posts are classified into $9$ narratives based on the evaluation of the content.

\textbf{External Signal Data from Google Trends.}
We collected Google Trends data to provide external signals aligned with both the temporal and thematic scope of the data sets.
Keywords—selected via frequency analysis of post content and expert consultation—were theme-specific for SocialSense (e.g.,\ ``bushfire evacuation,'' ``climate hoax'') and narrative-specific for DiN (e.g.,\ ``election fraud,'' ``deep state'').
Using the Google Trends API~\footnote{\url{https://serpapi.com/google-trends-api}}, the search information was retrieved via key phrases, normalized to 0--100, and retrieved at 10-minute intervals across the full temporal range of the social media datasets. These data were processed in an external signal timeline $G = {(t_k, g_k)}_{k=1}^{l}$, where $g_k$ denotes the search interest at time $t_k$, and aligned with each post using a temporal lag window $\tau{\text{lag}}$ to model potential causal influences on engagement.

\subsection{Models and Baselines}
We evaluate eight architectural variants, four Transformer-based and four Mamba-based, each modified to incorporate external signals, with details in \cref{app:model_details}.

\textbf{Transformer-based Variants.}
We extend the Transformer~\citep{Vaswani+2017} to model external signals via: \textit{+Token} embeds tokenized signals with engagement data for joint self-attention; \textit{+Attention} adds attention heads with custom masks for signal dependencies; \textit{+Layer} processes signals via MLP, integrating with sequences through cross-attention; \textit{+Adapter} inserts lightweight adapters~\citep{houlsby2019parameter} to model external signals.

\textbf{Mamba-based Variants.}
We adapt Mamba~\citep{gu2024mamba}, a selective state space model, to incorporate signals via: \textit{+Token} embeds tokenized signals with engagement data for joint processing; \textit{+Selection} conditions the selective scan on signal-engagement temporal relationships; \textit{+Layer} processes signals via MLP, integrating outputs into Mamba’s state; \textit{+Adapter} adds adapters to condition state transitions on signal intensities.

We compare against three baselines: vanilla TimeSeries Transformer, Mamba and MBPP~\citep{rizoiu2022interval}, a marked point process that models engagement as a self-exciting process.

\subsection{Results and Analysis}
\label{subsec:results}

\textbf{RQ1: Joint Modeling and Predictive Accuracy.}
\cref{tab:rmse_cf_comparison} compares the performance of architectural variants under counterfactual perturbations (see scenarios of \cref{subsubsec:counterfactual}).
\textit{Mamba+Adapter} achieves the lowest RMSE ($0.113$) on baseline data, with selective state space models effectively capturing temporal engagement dependencies. 
As exposure levels increase, RMSE rises non-linearly, suggesting higher exposure renders engagement patterns more predictable through stronger signal regularization. 
Experiments reveal a 1-3 day optimal intervention window, with Mamba-based models handling temporally distant counterfactuals better than Transformer variants. 
Extended policy durations challenge all models, though \textit{Transformer+Adapter} shows greater resilience to sustained signals. 
\cref{app:results} reveals disinformation (DiN dataset) poses greater challenges than misinformation, with higher base RMSE and larger increases at $60\%$ exposure, suggesting inherently greater difficulty in predicting engagement with deliberately deceptive content.

\begin{table}[t]
    \centering
    \caption{Root Mean Squared Error (RMSE) of engagement prediction over a 7-day horizon under counterfactual scenarios (see \cref{subsubsec:counterfactual}), averaged across two datasets. \textbf{Base}: Full exposure to external signal. \textbf{Scenario 1}: varying exposure levels (20\%, 40\%, 60\%); \textbf{Scenario 2}: altered treatment timing (-5, -3, -1 days); \textbf{Scenario 3}: different treatment durations (1, 3, 5 days). Lower RMSE indicates better performance. `*' denotes no external signals.}
    \label{tab:rmse_cf_comparison}
    \begin{adjustbox}{max width=\linewidth}
        \begin{tabular}{lcccccccccc}
            \toprule
            \toprule
            \textbf{Model}                   & {\textbf{Base}} & \multicolumn{3}{c}{\textbf{Scenario 1: Exposure}} & \multicolumn{3}{c}{\textbf{Scenario 2: Timing }} & \multicolumn{3}{c}{\textbf{Scenario 3: Duration}}                                                                                                       \\
                                             &                 & 20\%                                              & 40\%                                             & 60\%                                              & -5 days        & -3 days        & -1 day         & 1-day          & 3-day          & 5-day          \\
            \midrule
            Transformer~\citep{Vaswani+2017} & 0.193*          & --                                                & --                                               & --                                                & --             & --             & --             & --             & --             & --             \\
            Mamba~\citep{gu2024mamba}        & 0.189*          & --                                                & --                                               & --                                                & --             & --             & --             & --             & --             & --             \\
            MBPP~\citep{rizoiu2022interval}  & 0.193           & 0.326                                             & 0.295                                            & 0.281                                             & 0.225          & 0.225          & 0.226          & 0.329          & 0.315          & 0.284          \\ \midrule
            Transformer + Token              & 0.128           & 0.214                                             & 0.220                                            & 0.225                                             & 0.210          & 0.203          & 0.197          & 0.238          & 0.245          & 0.250          \\
            Transformer + Attention          & 0.122           & 0.198                                             & 0.191                                            & 0.183                                             & 0.194          & 0.188          & 0.182          & 0.224          & 0.198          & 0.194          \\
            Transformer + Layer              & 0.125           & 0.202                                             & 0.194                                            & 0.187                                             & 0.198          & 0.192          & 0.186          & 0.228          & 0.202          & 0.198          \\
            Transformer + Adapter            & 0.115           & 0.190                                             & 0.183                                            & 0.176                                             & 0.186          & 0.180          & 0.175          & \textbf{0.216} & \textbf{0.190} & \textbf{0.186} \\
            Mamba + Token                    & 0.129           & 0.205                                             & 0.197                                            & 0.190                                             & 0.201          & 0.195          & 0.189          & 0.245          & 0.225          & 0.218          \\
            Mamba + Selection                & 0.118           & 0.193                                             & 0.186                                            & 0.179                                             & 0.189          & 0.183          & 0.178          & 0.240          & 0.220          & 0.213          \\
            Mamba + Layer                    & 0.120           & 0.195                                             & 0.188                                            & 0.180                                             & 0.191          & 0.185          & 0.179          & 0.242          & 0.222          & 0.215          \\
            Mamba + Adapter                  & \textbf{0.113}  & \textbf{0.185}                                    & \textbf{0.178}                                   & \textbf{0.171}                                    & \textbf{0.181} & \textbf{0.176} & \textbf{0.170} & 0.236          & 0.218          & 0.210          \\
            \bottomrule
            \bottomrule
        \end{tabular}
    \end{adjustbox}
\end{table}
\begin{table}[t]
\centering
\caption{Binary Cross Entropy (BCE) evaluates treatment intensity prediction accuracy over a 7-day horizon across counterfactual scenarios, averaged across two datasets. Lower BCE reflects better modeling of temporal external signal patterns.}
\label{tab:bce_joint_modeling}
\begin{adjustbox}{max width=\linewidth}
\begin{tabular}{l|ccc|ccc|ccc}
\toprule
\toprule
\textbf{Model} & \multicolumn{3}{c|}{\textbf{Scenario 1: Exposure}} & \multicolumn{3}{c|}{\textbf{Scenario 2: Timing}} & \multicolumn{3}{c}{\textbf{Scenario 3: Treatment}} \\
 & 20\% & 40\% & 60\% & -5 days & -3 days & -1 day & 1-day & 3-day & 5-day \\
\midrule
Transformer + Token & 0.543 & 0.465 & 0.405 & 0.482 & 0.433 & 0.408 & 0.542 & 0.481 & 0.463 \\
Transformer + Attention & 0.510 & 0.421 & 0.378 & 0.450 & 0.390 & 0.352 & 0.533 & 0.465 & 0.446 \\
Transformer + Layer & 0.518 & 0.432 & 0.385 & 0.460 & 0.403 & 0.372 & \textbf{0.513} & \textbf{0.430} & \textbf{0.412} \\
Transformer + Adapter & 0.495 & 0.402 & 0.355 & 0.428 & 0.372 & 0.339 & 0.545 & 0.482 & 0.464 \\
Mamba + Token & 0.532 & 0.455 & 0.395 & 0.470 & 0.420 & 0.400 & 0.535 & 0.482 & 0.464 \\
Mamba + Selection & 0.502 & 0.413 & 0.368 & 0.435 & 0.378 & 0.346 & 0.535 & 0.470 & 0.454 \\
Mamba + Layer & 0.509 & 0.422 & 0.372 & 0.445 & 0.386 & 0.355 & 0.538 & 0.472 & 0.450 \\
Mamba + Adapter & \textbf{0.488} & \textbf{0.398} & \textbf{0.350} & \textbf{0.420} & \textbf{0.365} & \textbf{0.335} & 0.525 & 0.465 & 0.440 \\
\bottomrule
\bottomrule
\end{tabular}%
\end{adjustbox}
\end{table}
\begin{table}[t]
\centering
\caption{Average Treatment Effect (ATE), computed via G-computation~\citep{robins1986new}, quantifies causal impacts of interventions over a 7-day horizon with 95\% bootstrap confidence intervals ($\pm$), normalized across four engagement metrics (likes, comments, emoji reactions, shares). Higher ATE indicates stronger effects.}
\label{tab:ate_results}
\begin{adjustbox}{max width=\linewidth}
\begin{tabular}{l|ccc|ccc|ccc}
\toprule
\toprule
\multirow{2}{*}{\textbf{Model}} 
& \multicolumn{3}{c|}{\textbf{Exposure Adjustment}} 
& \multicolumn{3}{c|}{\textbf{Timing Adjustment}} 
& \multicolumn{3}{c}{\textbf{Duration Adjustment}} \\
\cmidrule(lr){2-4} \cmidrule(lr){5-7} \cmidrule(lr){8-10}
& \makecell{\textbf{CF1-1} \\ (0\%~$\rightarrow$~20\%)} 
& \makecell{\textbf{CF1-2} \\ (20\%~$\rightarrow$~40\%)} 
& \makecell{\textbf{CF1-3} \\ (40\%~$\rightarrow$~60\%)} 
& \makecell{\textbf{CF2-1} \\ (5-day early, day 0)} 
& \makecell{\textbf{CF2-2} \\ (3-day early, day 0)} 
& \makecell{\textbf{CF2-3} \\ (1-day early, day 0)} 
& \makecell{\textbf{CF3-1} \\ (1-day dur.)} 
& \makecell{\textbf{CF3-2} \\ (3-day dur.)} 
& \makecell{\textbf{CF3-3} \\ (5-day dur.)} \\
\midrule
Transformer+Token 
& 0.173  $\pm$ 0.109 & 0.334  $\pm$ 0.181 & 0.496  $\pm$ 0.102 
& 0.146  $\pm$ 0.075 & 0.188  $\pm$ 0.060 & 0.221  $\pm$ 0.054 
& 0.162  $\pm$ 0.046 & 0.297  $\pm$ 0.086 & 0.425  $\pm$ 0.104 \\
\midrule
Transformer+Attention 
& 0.180  $\pm$ 0.050 & 0.349  $\pm$ 0.083 & 0.523  $\pm$ 0.111 
& 0.158  $\pm$ 0.083 & 0.202  $\pm$ 0.067 & 0.245  $\pm$ 0.057 
& 0.172  $\pm$ 0.052 & 0.312  $\pm$ 0.090 & 0.456  $\pm$ 0.111 \\
\midrule
Transformer+Layer 
& 0.161  $\pm$ 0.047 & 0.297  $\pm$ 0.075 & 0.443  $\pm$ 0.091 
& 0.131  $\pm$ 0.067 & 0.170  $\pm$ 0.055 & 0.203  $\pm$ 0.048 
& 0.148  $\pm$ 0.045 & 0.269  $\pm$ 0.076 & 0.389  $\pm$ 0.094 \\
\midrule
Transformer+Adapter 
& 0.189  $\pm$ 0.052 & 0.364  $\pm$ 0.088 & 0.547  $\pm$ 0.117
& 0.166  $\pm$ 0.085 & 0.217  $\pm$ 0.071 & 0.257  $\pm$ 0.061 
& 0.184  $\pm$ 0.056 & 0.334  $\pm$ 0.097 & 0.479  $\pm$ 0.117 \\
\midrule
Mamba+Token 
& 0.197  $\pm$ 0.054 & 0.381  $\pm$ 0.091 & 0.574  $\pm$ 0.121 
& 0.175  $\pm$ 0.090 & 0.230  $\pm$ 0.075 & 0.272  $\pm$ 0.065 
& 0.192  $\pm$ 0.060 & 0.352  $\pm$ 0.102 & 0.503  $\pm$ 0.123 \\
\midrule
Mamba+Selection 
&\textbf{0.221  $\pm$ 0.060} & \textbf{0.425  $\pm$ 0.101} & 0.635  $\pm$ 0.135 
& \textbf{0.195  $\pm$ 0.099} & \textbf{0.256  $\pm$ 0.084} & \textbf{0.301  $\pm$ 0.074} 
& \textbf{0.216  $\pm$ 0.067} & \textbf{0.391  $\pm$ 0.112} & \textbf{0.559  $\pm$ 0.134} \\
\midrule
Mamba+Layer 
& 0.193  $\pm$ 0.052 & 0.371  $\pm$ 0.089 & 0.556  $\pm$ 0.118 
& 0.170  $\pm$ 0.087 & 0.224  $\pm$ 0.073 & 0.264  $\pm$ 0.063 
& 0.187  $\pm$ 0.058 & 0.340  $\pm$ 0.098 & 0.487  $\pm$ 0.119 \\
\midrule
Mamba+Adapter 
& 0.212  $\pm$ 0.058 & 0.409  $\pm$ 0.098 & \textbf{0.612  $\pm$ 0.131} 
& 0.187  $\pm$ 0.096 & 0.246  $\pm$ 0.080 & 0.290  $\pm$ 0.069 
& 0.208  $\pm$ 0.064 & 0.378  $\pm$ 0.109 & 0.538  $\pm$ 0.131 \\
\bottomrule
\bottomrule
\end{tabular}%
\end{adjustbox}
\end{table}

\textbf{RQ2: Architectural Differences in Temporal Causal Dependency Modeling.}
As shown in \cref{tab:bce_joint_modeling}, \textit{Mamba+Adapter} outperforms other architectures in capturing treatment dynamics, achieving the lowest BCE ($0.35$ at $60$\% exposure) with a $10.9$\% improvement over the best Transformer variant. An inverse relationship emerges between treatment intensity and modeling difficulty: BCE decreases from $0.488$ to $0.350$ as exposure rises from $20$\% to $60$\%. This suggests that high-intensity external signals may be more consistent with predictable patterns, offering potential for early detection of viral content trends.
Temporal misalignment further amplifies architectural differences. For 5-day early interventions, Mamba+Adapter demonstrates a $20.2$\% BCE advantage over Transformer+Token, showing the strength of state space models in managing temporally shifted causal processes.

\textbf{RQ3: Counterfactual Robustness and Reliability.}
\cref{tab:ate_results} reveals distinct patterns in causal relationships between external signals and engagement outcomes. The Mamba+Selection configuration estimates the highest ATE ($0.635$ $\pm$ $0.135$ for CF1-3, $40$\%~$\rightarrow$~$60$\% exposure), surpassing alternatives by $16.1$\%. This benefits from its selective scan algorithm, which is good at isolating causally relevant temporal patterns. Exposure adjustment experiments indicate a super-linear relationship between signal intensity and engagement impact: ATE rises by $42.1$\% from $20$\%~$\rightarrow$~$40$\% ($0.409$) to $40$\%~$\rightarrow$~$60$\% ($0.612$) exposure levels, suggesting that concentrated bursts of attention yield disproportionately high engagement returns.
Timing manipulations reveal a temporal window amplifying causal influence, with an average ATE ratio of $1.55$ between 1-day and 5-day early interventions across models, suggesting that precise timing within narrower windows boosts engagement, informing optimal content scheduling and promotion strategies.
Having identified the optimal architectural choice---Mamba+Adaptors, dubbed \emph{causal-Mamba}---, we now turn to identifying causal influencers.

\section{Case Study: Identifying Causal Influencers on Social Media}
\begin{figure}[t]
\begin{center}
\includegraphics[width=\linewidth]{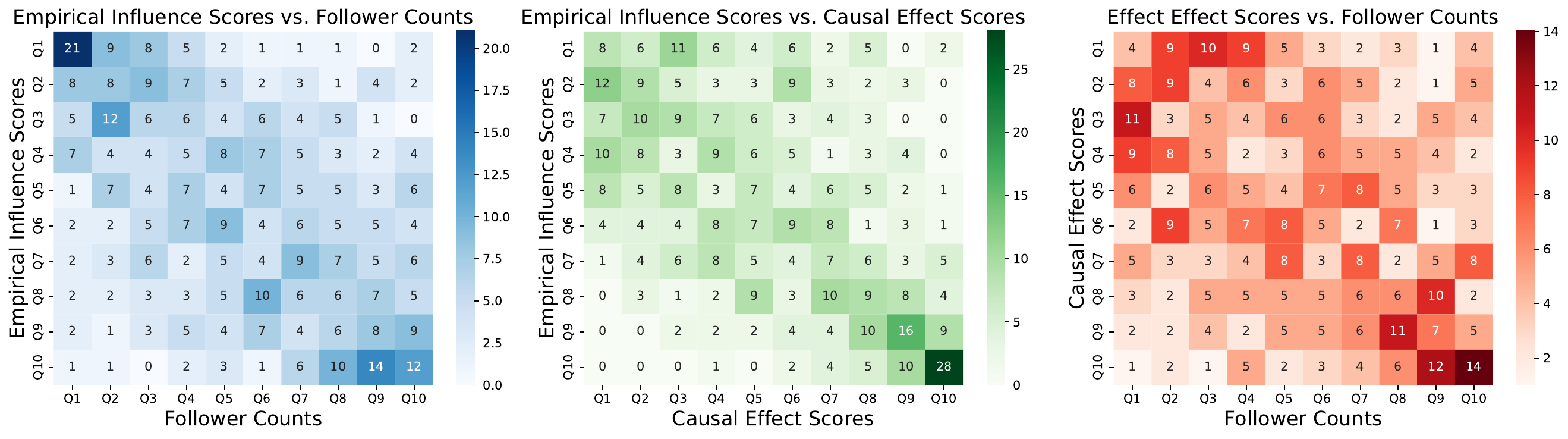}
\end{center}
\caption{
    Decile Heatmaps (Spearman $\rho$ correlation coefficient, Kendall's $W$ rank agreement, Concordance Correlation Coefficient $\mathrm{CCC}$): 
    \textbf{(left)} Follower Counts vs.\ Empirical Influence ($\rho = 0.49$, $W = 0.67$, $\mathrm{CCC} = 0.00$), 
    \textbf{(centre)} Causal Effect vs.\ Empirical Influence ($ 0.57$, $ 0.70$, $ 0.21$), 
    \textbf{(right)} Follower Counts vs.\ Causal Effect ($ 0.32$, $ 0.21$, $ 0.01$).
}
\label{fig:casestudy_heatmap}
\end{figure}

\textbf{Causal effect influence.}
We use our causal-Mamba architecture to estimate the causal influence of a source -- users, public page and groups.
We use the source's posting behavior as the causal treatment (compared to no exogenous signal), and we estimate its impact on predicting engagement metrics (quotes, replies, retweets, favorites).

\textbf{Baselines and datasets.}
There exists no universally accepted ground truth for measuring influence on social media.
A popular proxy is the \emph{number of followers}, as highly followed users are more likely to spread a content, possible leading to adoption (therefore, influence).
The closest to a gold standard is the perceived \emph{empirical influence} measure
\citep{ram2024empirically}, based on pairwise comparisons and an active learning-based ranking model.
The empirical influence measure is based on human judgements, and has been shown to have desirable psychosocial properties.
It is however prohibitive to scale; \citet{ram2024empirically} only computed it on $492$ X/Twitter users, discussing anti-climate change topics.
% containing the empirical influence scores of $492$ users. 

% To assess our joint causal framework, we applied the causal-Mamba architecture to two datasets. Note that there is no universally accepted ground truth for measuring individual influence on social networks, as influence is a multifaceted and context-dependent concept. Therefore, we rely on multiple proxy methods to estimate the influence of specific users. These include empirical influence scores derived from observed interactions, such as likes, shares, or retweets, and predicted causal-effect scores generated by our model, which attempt to quantify a user’s impact on others’ behaviors.

% First, we analyzed the \texttt{\#ArsonEmergency} dataset from X~\citep{ram2024empirically}, with validated influence scores for $492$ users. 

\textbf{Causal effect is a tighter approximation of influence.}
\cref{fig:casestudy_heatmap} shows pairs of Spearman correlation between the gold standard empirical influence and two approximations: the follower count and our proposed causal effect influence.
Visibly, the causal effect is better estimation for empirical influence ($\rho=0.57$) than follower counts ($\rho=0.49$), particularly visible for the top $10\%$ most influential users.
Further analysis using Kendall's W confirms this stronger rank agreement between causal effect and empirical influence ($W=0.70$) compared to followers ($W=0.67$). 
The Concordance Correlation Coefficient reveals that while follower count fails entirely to capture the magnitude of influence ($CCC=0.00$), causal effect maintains some concordance ($CCC=0.21$). 
In particular, the two approximation measures themselves show minimal agreement ($\rho=0.32$, $W=0.21$, $CCC=0.01$), suggesting they capture fundamentally different aspects of influence. 
These results challenge the common assumption that account popularity (follower count) is a reliable approximation for true influence, demonstrating causal effect as a superior estimate both in ranking and scale.

\textbf{Identify influential sources of misinformation.}
\begin{figure}[t]
\begin{center}
\includegraphics[width=0.8\linewidth]{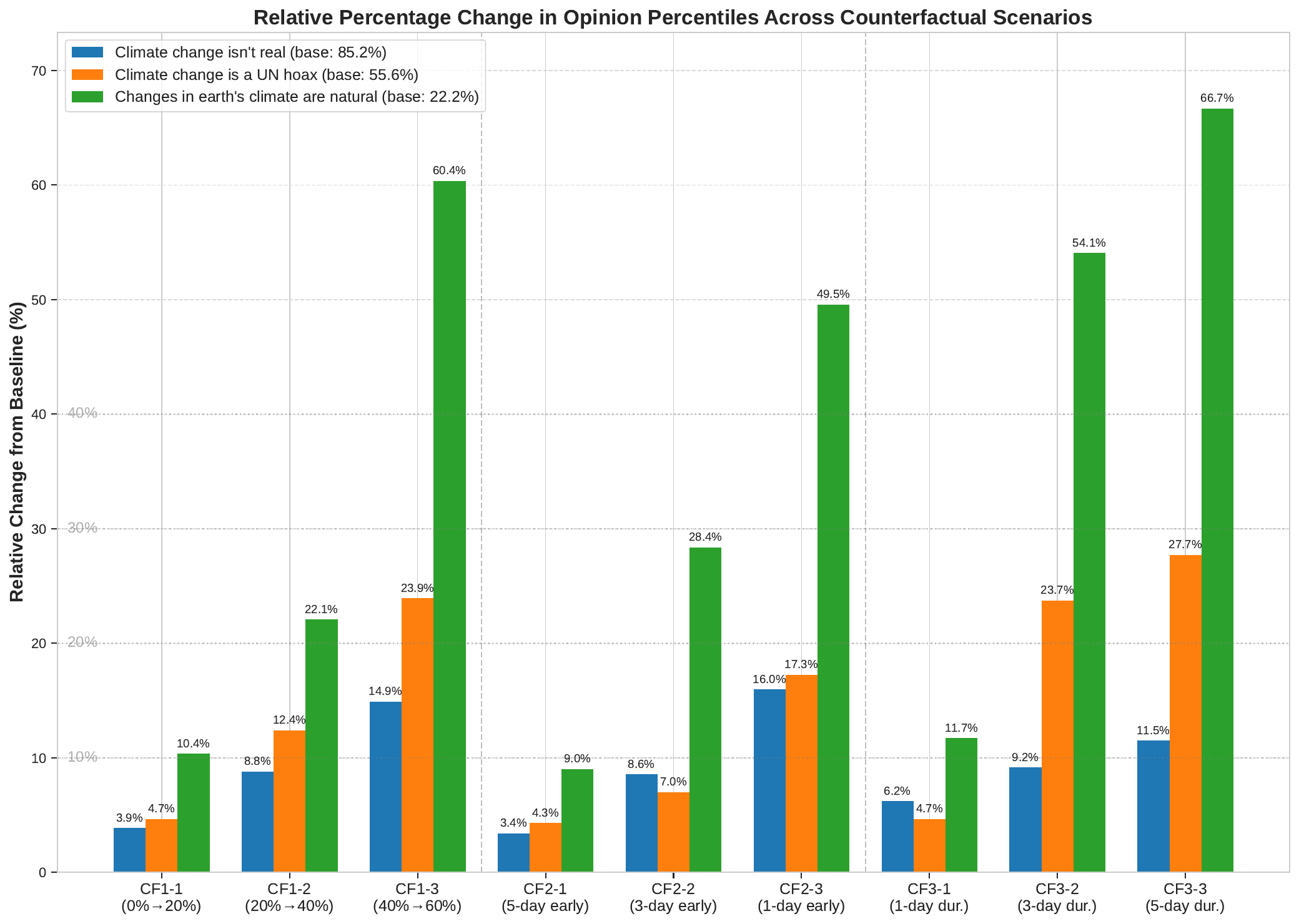}
\end{center}
\caption{
Relative percentage changes in engagement for three climate change misinformation narratives under counterfactual scenarios.
}
\label{fig:casestudy_heatmap}
\end{figure}

We analyzed the SocialSense dataset~\citep{kong2022slipping}, focusing specifically on vaccination and climate change narratives. 
For each opinion, we calculated a weighted composite engagement score: 
$E(o) = \hat{e}_{\text{likes}}(o) + \hat{e}_{\text{shares}}(o) + 3 \times \hat{e}_{\text{comments}}(o) + \hat{e}_{\text{emoji}}(o)$, 
then normalized these scores using percentile ranking:
$P(o) = \frac{\text{number of opinions with } E \leq E(o)}{\text{total number of opinions}}$. 

Our analysis extends viral potential theory~\cite{rizoiu2017online} to misinformation contexts, confirming that content responsiveness to external promotion depends on baseline popularity and temporal dynamics across three climate change narratives: ``Climate change isn't real'' ($85.2$th percentile), ``Climate change is a UN hoax'' ($55.6$th percentile), and ``Changes in earth's climate are natural'' ($22.2$th percentile). Under progressive exposure manipulation (CF1), the lowest-baseline narrative exhibits super-linear amplification with a 6-fold scaling coefficient ($10.4$\%~$\rightarrow$~$22.1\%$~$\rightarrow$~$60.4$\%), confirming that small external signals can trigger massive engagement increases. This behavior validates high-potential, low-baseline content theory~\citep{rizoiu2017online}, showing that emerging misinformation operates similarly to sleeping beauty content with high viral potential that have yet to achieve widespread attention but can rapidly amplify under external promotion.
The intermediate narrative shows moderate scaling effects ($4.7$\%~$\rightarrow$~$12.4$\%~$\rightarrow$~$23.9$\%), consistent with non-linear threshold dynamics in social influence, as evidenced by tipping point experiments~\citep{centola2018experimental}.
Temporal intervention analysis (CF2) reveals that initiating promotion one day earlier yields the highest amplification across all narrative types ($16.0$\%, $17.3$\%, and $28.4$\%, respectively). This finding aligns with prior work on optimal promotion timing~\citep{rizoiu2017online} and supports the broader framework of social acceleration~\citep{rosa2013social}.
Extended-duration experiments (CF3) show that a sustained 5-day exposure maximizes narrative amplification, with the emerging narrative reaching a $66.7$\% increase. This result corroborates the mere exposure effect~\citep{zajonc1968attitudinal}, reinforcing the superiority of prolonged campaigns over short-term bursts.

Our causal model further identified influential public groups that amplified misinformation. For instance, @AustraliansforSafeTechnology increased the engagement score for ``5G/smart tech is unsafe'' narratives from $0.51$ to $0.63$ in February 2020, while @ClimateChangeBattleRoyale elevated ``Climate change crisis isn't real'' content from $0.11$ to $0.23$ in September 2019.

\section{Conclusion}

We investigate how external signals drive social media engagement, a key factor in misinformation and disinformation spread, using a straightforward joint treatment-outcome approach. 
By tweaking advanced models such as Transformers and Mamba, we improve the predictions of engagement under realistic policy treatments. 
Our approach faces \emph{inherent causal challenges} including unobserved confounding from algorithmic amplification, selection bias in user engagement patterns, and treatment interference through network effects. 
Additional \emph{limitations} include dependence on high-quality external signals that aren't always available, sensitivity to rapidly evolving platform algorithms, and assumptions of temporal stability in user behavior patterns. 
Despite these constraints, our case study proves our causal effect measure aligns closely with empirical influence, offering platforms and regulators a reliable metric to identify influential spreaders of misleading narratives.

\section*{Ethical Considerations and Dual-Use Concerns}
% This work uses publicly available social media data from platforms like X (formerly Twitter) and Facebook to analyze engagement patterns and misinformation dynamics. All data were collected in accordance with platform terms of service and API usage policies, and with approvals from our institution's Human Ethics Committee. Our analysis examines aggregate trends at the group level rather than targeting individuals. User identifiers were anonymized, and no personally identifiable information was collected or retained.
% While the framework developed could have applications in marketing or content strategy, its primary purpose is to enhance scientific understanding of engagement and the spread of misinformation.
% % , not to target specific narratives or users for manipulative purposes.

This work uses publicly available social media data from platforms like X (formerly Twitter) and Facebook to analyze engagement patterns and misinformation dynamics. All data were collected in accordance with platform terms of service and API usage policies, and with approvals from our institution's Human Ethics Committee. Our analysis examines aggregate trends at the group level rather than targeting individuals. User identifiers were anonymized, and no personally identifiable information was collected or retained.

Our causal influence estimation framework has important ethical implications that extend beyond data privacy. While these tools enable beneficial applications like identifying sources of misinformation and improving platform governance, they could potentially be misused to optimize manipulation campaigns or target influential users for spreading harmful content. We acknowledge these dual-use concerns and propose safeguards, such as
% : (1) implementing transparent reporting requirements when deploying influence measurement tools, (2) establishing multi-stakeholder oversight when applying these methods at scale, and (3) 
integrating with existing content moderation frameworks to prioritize intervention efforts.

We believe that understanding the causal mechanisms of influence is necessary for designing effective interventions against harmful content, and that responsible development practices—including limitations on individual-level influence measurements—can help mitigate potential misuse. Our research aims to contribute to the broader effort of creating healthier online information ecosystems while remaining mindful of these ethical challenges.

\bibliography{colm2025_conference}
\bibliographystyle{colm2025_conference}

\appendix
\section{Appendix}
\subsection{Related Work}

Our research lies at the intersection of social media engagement prediction, causal inference for sequential data, and advanced sequence modeling. 

\paragraph{Social Media Engagement Prediction.}
Social media engagement prediction has evolved from early content and user-focused models \citep{cheng2014can} to approaches incorporating network structures \citep{zhao2015seismic} and temporal dynamics \citep{rizoiu2017expecting}.
Recent deep learning advances include DeepCas \citep{li2017deepcas}, which uses random walks and attention mechanisms for cascade prediction; SNPP \citep{ding2019social}, applying temporal point processes with recurrent architectures; and Topo-LSTM \citep{wang2017topological}, which models user interactions through topological structures. Graph-based approaches like DeepInf \citep{qiu2018deepinf} leverage network structures to predict user behaviors.
Temporal dynamics have been addressed through several specialized frameworks. DeepHawkes \citep{cao2017deephawkes} combines reinforcement learning with Hawkes processes, while HIP \citep{rizoiu2017expecting} and MBPP \citep{rizoiu2022interval} model view count dynamics on YouTube. IC-TH \citep{kong2023interval} tackles interval-censored observations in retweet prediction, and OMM \citep{calderon2024opinion} offers a cross-platform mathematical framework for predicting content spread.

While these approaches have cascade modeling, they typically rely on historical engagement patterns without considering external causal drivers. Our work explicitly models how external signals (like Google Trends) causally impact engagement dynamics.

\paragraph{Causal Inference for Sequential Data.}
Causal inference with sequential data builds on foundations like g-methods \citep{robins1986new} and marginal structural models \citep{robins2000marginal}, which handle time-varying treatments and confounders. These approaches have been extended to continuous-time settings \citep{lok2008statistical, schulam2017reliable}.
Neural approaches include Recurrent Marginal Structural Networks \citep{lim2018forecasting}, which extend traditional causal models with recurrent architectures, and Counterfactual Recurrent Network \citep{Bica2020Estimating}, which uses adversarial learning to address time-varying confounding.
For continuous-time treatments, researchers have developed personalized treatment response curves using Gaussian processes \citep{soleimani2017treatment} and counterfactual frameworks for disease trajectory modeling \citep{schulam2017reliable}. Recently, \citet{hizli2023causal} introduced a joint treatment-outcome model combining marked point processes with Gaussian processes for healthcare policy interventions.
These methods, while advanced, often lack the flexibility to capture social media's unique engagement dynamics and the complex temporal patterns of external influence.

\paragraph{State Space Models and Transformers for Sequence Modeling.}
Sequence modeling has recently been dominated by transformer architectures \citep{Vaswani+2017} and state space models. Time series adaptations include Informer \citep{zhou2021informer}, using ProbSparse self-attention, and Autoformer \citep{wu2021autoformer}, leveraging auto-correlation for period-based dependencies.
State Space Models (SSMs) have emerged as efficient alternatives for capturing long-range dependencies \citep{gu2020hippo}, processing extremely long sequences \citep{dao2022flashattention}, and achieving competitive performance in language modeling \citep{gu2024mamba}. The Mamba architecture \citep{gu2024mamba} introduced selective state space modeling, achieving state-of-the-art results with reduced computational requirements.

Despite these advances, few studies have applied modern SSMs to social media engagement prediction in causal contexts. Existing approaches typically rely on graph-based \citep{lu2023continuous}, RNN \citep{wang2017cascade}, or transformer methods \citep{zuo2020transformer} that assume uniform sampling or discrete snapshots, overlooking the fine-grained temporal patterns essential to engagement dynamics.
\subsection{Training Details}
\label{app:training_details}
Models are trained on 4x NVIDIA A100 GPUs with PyTorch, using the joint loss function from \cref{subsec:training} (\(\alpha = 0.5\), \(\beta = 0.1\)). Optimization uses AdamW with a learning rate of \(10^{-4}\), batch size of 64, and early stopping based on validation RMSE (patience = 10 epochs). Hyperparameters (e.g.,\ \(\tau_{\text{lag}} = 24\) hours, \(w = 6\)) are tuned via grid search.

\subsection{Model Architecture Details}
\label{app:model_details}
Our implementation uses a depth-$L$ architecture with embedding dimension $d$ and hidden dimension $h$. Here are the details for each integration mechanism:

\textbf{Token Integration:} For the token-based approach, external signals $g_t$ at time $t$ are embedded through a projection $\mathbf{E}_g \in \mathbb{R}^{|g| \times d}$:
\begin{equation*}
\mathbf{e}_g^t = \text{Embedding}(g_t) = \mathbf{W}_g g_t + \mathbf{b}_g
\end{equation*}

These embedded tokens are then interleaved with engagement embeddings in the input sequence:
\begin{equation*}
\mathbf{X} = [\mathbf{e}_1, \mathbf{e}_g^1, \mathbf{e}_2, \mathbf{e}_g^2, \ldots, \mathbf{e}_n, \mathbf{e}_g^n]
\end{equation*}

\textbf{Attention Integration:} For attention-based integration, we modify the self-attention mechanism to include specialized heads focused on learning external signals. Given query $\mathbf{Q}$, key $\mathbf{K}$, and value $\mathbf{V}$ matrices, we compute:
\begin{equation*}
\text{Attention}(\mathbf{Q}, \mathbf{K}, \mathbf{V}) = \text{softmax}\left(\frac{\mathbf{Q}\mathbf{K}^T}{\sqrt{d_k}} \odot \mathbf{M}\right)\mathbf{V}
\end{equation*}
where $\mathbf{M}$ is a masking matrix with elements $m_{ij} = \exp(-\beta|t_i - t_j|)$ that emphasizes temporally proximate signal-engagement pairs with decay parameter $\beta = 0.1$.

\textbf{Layer Integration:} Our layer-based approach processes external signals through a separate MLP layer:
\begin{equation*}
\mathbf{h}_g^l = \text{MLP}_g^l(\mathbf{h}_g^{l-1}) = \mathbf{W}_2^l \cdot \text{GeLU}(\mathbf{W}_1^l \mathbf{h}_g^{l-1} + \mathbf{b}_1^l) + \mathbf{b}_2^l
\end{equation*}
which is merged with the main sequence representations through cross-attention at each layer $l$:
\begin{equation*}
\mathbf{h}_e^l = \mathbf{h}_e^{l-1} + \text{CrossAttention}(\mathbf{h}_e^{l-1}, \mathbf{h}_g^l, \mathbf{h}_g^l)
\end{equation*}

\textbf{Adapter Integration:} For adapter-based integration, we insert compact bottleneck modules after each self-attention and feed-forward layer:
\begin{equation*}
\mathbf{h}_{\text{out}} = \mathbf{h}_{\text{in}} + \mathbf{W}_{\text{up}} \cdot \text{GeLU}(\mathbf{W}_{\text{down}} \cdot \mathbf{h}_{\text{in}} + \mathbf{b}_{\text{down}}) + \mathbf{b}_{\text{up}}
\end{equation*}
where $\mathbf{W}_{\text{down}} \in \mathbb{R}^{d \times r}$ and $\mathbf{W}_{\text{up}} \in \mathbb{R}^{r \times d}$ with bottleneck dimension $r = d/8$.

For the Mamba variants, we extend the selective state space model with signal-conditioned selection mechanisms. The state update equation* becomes:
\begin{equation*}
\mathbf{h}_t = \mathbf{A}(\mathbf{g}_t)\mathbf{h}_{t-1} + \mathbf{B}(\mathbf{g}_t)\mathbf{x}_t
\end{equation*}
where the state transition matrix $\mathbf{A}(\mathbf{g}_t)$ and input projection $\mathbf{B}(\mathbf{g}_t)$ are conditioned on external signals $\mathbf{g}_t$ through a low-rank adaptation:
\begin{equation*}
\mathbf{A}(\mathbf{g}_t) = \mathbf{A}_0 + \Delta\mathbf{A} \cdot \sigma(\mathbf{W}_g \mathbf{g}_t)
\end{equation*}

The selective scan algorithm uses a similar signal-aware mechanism to compute selection weights, allowing the model to dynamically focus on relevant temporal dependencies based on external signal patterns.

\subsection{Detailed Results Across Individual Datasets}
\label{app:results}
This appendix presents detailed experimental results for the predictive performance of various model architectures across five distinct datasets: bushfire, covid, vaccination, climate change, and DiN. 
The tables provided here complement the aggregated results reported in \cref{subsec:results}, which represent the mean Root Mean Squared Error (RMSE) and Binary Cross Entropy (BCE) across these datasets. Here, we report individual dataset results to offer a granular view of model behavior under different counterfactual scenarios.

\begin{table}[t]
\centering
\begin{adjustbox}{max width=\linewidth}
\begin{tabular}{lcccccccccc}
\toprule
\toprule
\textbf{Model} & \textbf{Base} & \multicolumn{3}{c}{\textbf{Scenario 1: Exposure}} & \multicolumn{3}{c}{\textbf{Scenario 2: Timing}} & \multicolumn{3}{c}{\textbf{Scenario 3: Duration}} \\
 & & 20\% & 40\% & 60\% & -5 days & -3 days & -1 day & 1-day & 3-day & 5-day \\
\midrule
Transformer + Token & 0.147 & 0.233 & 0.242 & 0.243 & 0.232 & 0.213 & 0.207 & 0.252 & 0.263 & 0.272 \\
Transformer + Attention & 0.133 & 0.212 & 0.193 & 0.197 & 0.208 & 0.202 & 0.188 & 0.242 & 0.208 & 0.212 \\
Transformer + Layer & 0.142 & 0.208 & 0.192 & 0.193 & 0.212 & 0.198 & 0.192 & 0.238 & 0.212 & 0.208 \\
Transformer + Adapter & 0.123 & 0.192 & 0.173 & 0.182 & 0.188 & 0.182 & 0.173 & 0.222 & 0.188 & 0.192 \\
Mamba + Token & 0.143 & 0.207 & 0.183 & 0.192 & 0.203 & 0.197 & 0.183 & 0.263 & 0.247 & 0.232 \\
Mamba + Selection & 0.132 & 0.183 & 0.177 & 0.178 & 0.187 & 0.173 & 0.177 & 0.262 & 0.238 & 0.223 \\
Mamba + Layer & 0.128 & 0.177 & 0.162 & 0.158 & 0.173 & 0.167 & 0.153 & 0.252 & 0.228 & 0.217 \\
Mamba + Adapter & 0.127 & 0.183 & 0.172 & 0.173 & 0.187 & 0.183 & 0.172 & 0.258 & 0.232 & 0.208 \\
\bottomrule
\bottomrule
\end{tabular}
\end{adjustbox}
\caption{Root Mean Squared Error (RMSE) of predicted social media engagement metrics over a 7-day horizon under counterfactual scenarios for the DiN dataset.}
\label{tab:rmse_din}
\end{table}

\begin{table}[t]
\centering
\begin{adjustbox}{max width=\linewidth}
\begin{tabular}{l|ccc|ccc|ccc}
\toprule
\toprule
\textbf{Model} & \multicolumn{3}{c|}{\textbf{Scenario 1: Exposure}} & \multicolumn{3}{c|}{\textbf{Scenario 2: Timing}} & \multicolumn{3}{c}{\textbf{Scenario 3: Treatment}} \\
 & 20\% & 40\% & 60\% & -5 days & -3 days & -1 day & 1-day & 3-day & 5-day \\
\midrule
Transformer + Token & 0.592 & 0.502 & 0.438 & 0.527 & 0.458 & 0.437 & 0.583 & 0.522 & 0.488 \\
Transformer + Attention & 0.543 & 0.442 & 0.403 & 0.487 & 0.403 & 0.377 & 0.582 & 0.498 & 0.487 \\
Transformer + Layer & 0.563 & 0.457 & 0.417 & 0.493 & 0.432 & 0.393 & 0.562 & 0.463 & 0.437 \\
Transformer + Adapter & 0.532 & 0.423 & 0.387 & 0.453 & 0.397 & 0.358 & 0.578 & 0.507 & 0.493 \\
Mamba + Token & 0.577 & 0.488 & 0.437 & 0.503 & 0.457 & 0.433 & 0.572 & 0.513 & 0.497 \\
Mamba + Selection & 0.547 & 0.442 & 0.393 & 0.457 & 0.403 & 0.367 & 0.568 & 0.507 & 0.483 \\
Mamba + Layer & 0.552 & 0.443 & 0.397 & 0.463 & 0.407 & 0.368 & 0.572 & 0.497 & 0.483 \\
Mamba + Adapter & 0.533 & 0.427 & 0.378 & 0.452 & 0.388 & 0.362 & 0.558 & 0.493 & 0.477 \\
\bottomrule
\bottomrule
\end{tabular}
\end{adjustbox}
\caption{Binary Cross Entropy (BCE) of treatment intensity predictions over a 7-day horizon for the DiN dataset across counterfactual scenarios.}
\label{tab:bce_din}
\end{table}

\begin{table}[t]
\centering
\begin{adjustbox}{max width=\linewidth}
\begin{tabular}{lcccccccccc}
\toprule
\toprule
\textbf{Model} & \textbf{Base} & \multicolumn{3}{c}{\textbf{Scenario 1: Exposure}} & \multicolumn{3}{c}{\textbf{Scenario 2: Timing}} & \multicolumn{3}{c}{\textbf{Scenario 3: Duration}} \\
 & & 20\% & 40\% & 60\% & -5 days & -3 days & -1 day & 1-day & 3-day & 5-day \\
\midrule
Transformer + Token & 0.129 & 0.216 & 0.218 & 0.223 & 0.208 & 0.203 & 0.198 & 0.241 & 0.243 & 0.248 \\
Transformer + Attention & 0.123 & 0.198 & 0.193 & 0.183 & 0.193 & 0.188 & 0.183 & 0.223 & 0.198 & 0.193 \\
Transformer + Layer & 0.126 & 0.203 & 0.198 & 0.188 & 0.198 & 0.193 & 0.188 & 0.228 & 0.203 & 0.198 \\
Transformer + Adapter & 0.117 & 0.193 & 0.188 & 0.178 & 0.188 & 0.183 & 0.178 & 0.218 & 0.193 & 0.188 \\
Mamba + Token & 0.130 & 0.208 & 0.203 & 0.193 & 0.203 & 0.198 & 0.193 & 0.243 & 0.223 & 0.218 \\
Mamba + Selection & 0.119 & 0.198 & 0.193 & 0.183 & 0.193 & 0.188 & 0.183 & 0.238 & 0.218 & 0.213 \\
Mamba + Layer & 0.121 & 0.203 & 0.198 & 0.188 & 0.198 & 0.193 & 0.188 & 0.243 & 0.223 & 0.218 \\
Mamba + Adapter & 0.114 & 0.188 & 0.183 & 0.173 & 0.183 & 0.178 & 0.173 & 0.233 & 0.218 & 0.213 \\
\bottomrule
\bottomrule
\end{tabular}
\end{adjustbox}
\caption{Root Mean Squared Error (RMSE) of predicted social media engagement metrics over a 7-day horizon under counterfactual scenarios for the climate change dataset.}
\label{tab:rmse_climate}
\end{table}

\begin{table}[t]
\centering
\begin{adjustbox}{max width=\linewidth}
\begin{tabular}{l|ccc|ccc|ccc}
\toprule
\toprule
\textbf{Model} & \multicolumn{3}{c|}{\textbf{Scenario 1: Exposure}} & \multicolumn{3}{c|}{\textbf{Scenario 2: Timing}} & \multicolumn{3}{c}{\textbf{Scenario 3: Treatment}} \\
 & 20\% & 40\% & 60\% & -5 days & -3 days & -1 day & 1-day & 3-day & 5-day \\
\midrule
Transformer + Token & 0.538 & 0.463 & 0.403 & 0.478 & 0.433 & 0.408 & 0.538 & 0.478 & 0.463 \\
Transformer + Attention & 0.508 & 0.423 & 0.378 & 0.448 & 0.393 & 0.353 & 0.528 & 0.463 & 0.443 \\
Transformer + Layer & 0.513 & 0.433 & 0.383 & 0.458 & 0.403 & 0.373 & 0.508 & 0.428 & 0.413 \\
Transformer + Adapter & 0.493 & 0.403 & 0.353 & 0.428 & 0.373 & 0.338 & 0.543 & 0.483 & 0.463 \\
Mamba + Token & 0.528 & 0.453 & 0.393 & 0.468 & 0.418 & 0.398 & 0.533 & 0.478 & 0.463 \\
Mamba + Selection & 0.498 & 0.413 & 0.368 & 0.438 & 0.378 & 0.348 & 0.533 & 0.468 & 0.453 \\
Mamba + Layer & 0.503 & 0.423 & 0.373 & 0.448 & 0.388 & 0.358 & 0.538 & 0.473 & 0.448 \\
Mamba + Adapter & 0.483 & 0.398 & 0.348 & 0.418 & 0.363 & 0.333 & 0.523 & 0.458 & 0.438 \\
\bottomrule
\bottomrule
\end{tabular}
\end{adjustbox}
\caption{Binary Cross Entropy (BCE) of treatment intensity predictions over a 7-day horizon for the climate change dataset across counterfactual scenarios.}
\label{tab:bce_climate}
\end{table}

\begin{table}[t]
\centering
\begin{adjustbox}{max width=\linewidth}
\begin{tabular}{lcccccccccc}
\toprule
\toprule
\textbf{Model} & \textbf{Base} & \multicolumn{3}{c}{\textbf{Scenario 1: Exposure}} & \multicolumn{3}{c}{\textbf{Scenario 2: Timing}} & \multicolumn{3}{c}{\textbf{Scenario 3: Duration}} \\
 & & 20\% & 40\% & 60\% & -5 days & -3 days & -1 day & 1-day & 3-day & 5-day \\
\midrule
Transformer + Token & 0.126 & 0.211 & 0.213 & 0.218 & 0.203 & 0.198 & 0.193 & 0.235 & 0.238 & 0.243 \\
Transformer + Attention & 0.121 & 0.193 & 0.188 & 0.178 & 0.188 & 0.183 & 0.178 & 0.218 & 0.193 & 0.188 \\
Transformer + Layer & 0.123 & 0.198 & 0.193 & 0.183 & 0.193 & 0.188 & 0.183 & 0.223 & 0.198 & 0.193 \\
Transformer + Adapter & 0.114 & 0.188 & 0.183 & 0.173 & 0.183 & 0.178 & 0.173 & 0.213 & 0.188 & 0.183 \\
Mamba + Token & 0.127 & 0.203 & 0.198 & 0.188 & 0.198 & 0.193 & 0.188 & 0.238 & 0.218 & 0.213 \\
Mamba + Selection & 0.117 & 0.193 & 0.188 & 0.178 & 0.188 & 0.183 & 0.178 & 0.233 & 0.213 & 0.208 \\
Mamba + Layer & 0.119 & 0.198 & 0.193 & 0.183 & 0.193 & 0.188 & 0.183 & 0.238 & 0.218 & 0.213 \\
Mamba + Adapter & 0.112 & 0.183 & 0.178 & 0.168 & 0.178 & 0.173 & 0.168 & 0.228 & 0.213 & 0.208 \\
\bottomrule
\bottomrule
\end{tabular}
\end{adjustbox}
\caption{Root Mean Squared Error (RMSE) of predicted social media engagement metrics over a 7-day horizon under counterfactual scenarios for the vaccination dataset.}
\label{tab:rmse_vaccination}
\end{table}

\begin{table}[t]
\centering
\begin{adjustbox}{max width=\linewidth}
\begin{tabular}{l|ccc|ccc|ccc}
\toprule
\toprule
\textbf{Model} & \multicolumn{3}{c|}{\textbf{Scenario 1: Exposure}} & \multicolumn{3}{c|}{\textbf{Scenario 2: Timing}} & \multicolumn{3}{c}{\textbf{Scenario 3: Treatment}} \\
 & 20\% & 40\% & 60\% & -5 days & -3 days & -1 day & 1-day & 3-day & 5-day \\
\midrule
Transformer + Token & 0.533 & 0.458 & 0.398 & 0.473 & 0.428 & 0.403 & 0.533 & 0.473 & 0.458 \\
Transformer + Attention & 0.503 & 0.418 & 0.373 & 0.443 & 0.388 & 0.348 & 0.523 & 0.458 & 0.438 \\
Transformer + Layer & 0.508 & 0.428 & 0.378 & 0.453 & 0.398 & 0.368 & 0.503 & 0.423 & 0.408 \\
Transformer + Adapter & 0.488 & 0.398 & 0.348 & 0.423 & 0.368 & 0.333 & 0.538 & 0.478 & 0.458 \\
Mamba + Token & 0.523 & 0.448 & 0.388 & 0.463 & 0.413 & 0.393 & 0.528 & 0.473 & 0.458 \\
Mamba + Selection & 0.493 & 0.408 & 0.363 & 0.433 & 0.373 & 0.343 & 0.528 & 0.463 & 0.448 \\
Mamba + Layer & 0.498 & 0.418 & 0.368 & 0.443 & 0.383 & 0.353 & 0.533 & 0.468 & 0.443 \\
Mamba + Adapter & 0.478 & 0.393 & 0.343 & 0.413 & 0.358 & 0.328 & 0.518 & 0.453 & 0.433 \\
\bottomrule
\bottomrule
\end{tabular}
\end{adjustbox}
\caption{Binary Cross Entropy (BCE) of treatment intensity predictions over a 7-day horizon for the vaccination dataset across counterfactual scenarios.}
\label{tab:bce_vaccination}
\end{table}

\begin{table}[t]
\centering
\begin{adjustbox}{max width=\linewidth}
\begin{tabular}{lcccccccccc}
\toprule
\toprule
\textbf{Model} & \textbf{Base} & \multicolumn{3}{c}{\textbf{Scenario 1: Exposure}} & \multicolumn{3}{c}{\textbf{Scenario 2: Timing}} & \multicolumn{3}{c}{\textbf{Scenario 3: Duration}} \\
 & & 20\% & 40\% & 60\% & -5 days & -3 days & -1 day & 1-day & 3-day & 5-day \\
\midrule
Transformer + Token & 0.123 & 0.208 & 0.213 & 0.218 & 0.203 & 0.198 & 0.193 & 0.235 & 0.238 & 0.243 \\
Transformer + Attention & 0.117 & 0.193 & 0.188 & 0.178 & 0.188 & 0.183 & 0.178 & 0.218 & 0.193 & 0.188 \\
Transformer + Layer & 0.119 & 0.198 & 0.193 & 0.183 & 0.193 & 0.188 & 0.183 & 0.223 & 0.198 & 0.193 \\
Transformer + Adapter & 0.112 & 0.188 & 0.183 & 0.173 & 0.183 & 0.178 & 0.173 & 0.213 & 0.188 & 0.183 \\
Mamba + Token & 0.124 & 0.203 & 0.198 & 0.188 & 0.198 & 0.193 & 0.188 & 0.238 & 0.218 & 0.213 \\
Mamba + Selection & 0.115 & 0.193 & 0.188 & 0.178 & 0.188 & 0.183 & 0.178 & 0.233 & 0.213 & 0.208 \\
Mamba + Layer & 0.117 & 0.198 & 0.193 & 0.183 & 0.193 & 0.188 & 0.183 & 0.238 & 0.218 & 0.213 \\
Mamba + Adapter & 0.110 & 0.183 & 0.178 & 0.168 & 0.178 & 0.173 & 0.168 & 0.228 & 0.213 & 0.208 \\
\bottomrule
\bottomrule
\end{tabular}
\end{adjustbox}
\caption{Root Mean Squared Error (RMSE) of predicted social media engagement metrics over a 7-day horizon under counterfactual scenarios for the covid dataset.}
\label{tab:rmse_covid}
\end{table}

\begin{table}[t]
\centering
\begin{adjustbox}{max width=\linewidth}
\begin{tabular}{l|ccc|ccc|ccc}
\toprule
\toprule
\textbf{Model} & \multicolumn{3}{c|}{\textbf{Scenario 1: Exposure}} & \multicolumn{3}{c|}{\textbf{Scenario 2: Timing}} & \multicolumn{3}{c}{\textbf{Scenario 3: Treatment}} \\
 & 20\% & 40\% & 60\% & -5 days & -3 days & -1 day & 1-day & 3-day & 5-day \\
\midrule
Transformer + Token & 0.531 & 0.453 & 0.393 & 0.468 & 0.423 & 0.398 & 0.528 & 0.468 & 0.453 \\
Transformer + Attention & 0.498 & 0.413 & 0.368 & 0.438 & 0.383 & 0.343 & 0.518 & 0.453 & 0.433 \\
Transformer + Layer & 0.503 & 0.423 & 0.373 & 0.448 & 0.393 & 0.363 & 0.498 & 0.418 & 0.403 \\
Transformer + Adapter & 0.483 & 0.393 & 0.343 & 0.418 & 0.363 & 0.333 & 0.533 & 0.473 & 0.453 \\
Mamba + Token & 0.518 & 0.443 & 0.383 & 0.458 & 0.408 & 0.388 & 0.523 & 0.473 & 0.453 \\
Mamba + Selection & 0.488 & 0.403 & 0.358 & 0.428 & 0.368 & 0.338 & 0.523 & 0.458 & 0.443 \\
Mamba + Layer & 0.493 & 0.413 & 0.363 & 0.438 & 0.378 & 0.348 & 0.528 & 0.463 & 0.438 \\
Mamba + Adapter & 0.473 & 0.388 & 0.338 & 0.413 & 0.358 & 0.328 & 0.513 & 0.448 & 0.428 \\
\bottomrule
\bottomrule
\end{tabular}
\end{adjustbox}
\caption{Binary Cross Entropy (BCE) of treatment intensity predictions over a 7-day horizon for the covid dataset across counterfactual scenarios.}
\label{tab:bce_covid}
\end{table}

\begin{table}[t]
\centering
\begin{adjustbox}{max width=\linewidth}
\begin{tabular}{lcccccccccc}
\toprule
\toprule
\textbf{Model} & \textbf{Base} & \multicolumn{3}{c}{\textbf{Scenario 1: Exposure}} & \multicolumn{3}{c}{\textbf{Scenario 2: Timing}} & \multicolumn{3}{c}{\textbf{Scenario 3: Duration}} \\
 & & 20\% & 40\% & 60\% & -5 days & -3 days & -1 day & 1-day & 3-day & 5-day \\
\midrule
Transformer + Token & 0.121 & 0.206 & 0.212 & 0.217 & 0.202 & 0.193 & 0.188 & 0.231 & 0.237 & 0.242 \\
Transformer + Attention & 0.116 & 0.192 & 0.187 & 0.173 & 0.186 & 0.178 & 0.172 & 0.216 & 0.191 & 0.183 \\
Transformer + Layer & 0.117 & 0.196 & 0.188 & 0.181 & 0.192 & 0.187 & 0.182 & 0.218 & 0.193 & 0.192 \\
Transformer + Adapter & 0.111 & 0.183 & 0.182 & 0.171 & 0.179 & 0.173 & 0.168 & 0.211 & 0.186 & 0.178 \\
Mamba + Token & 0.123 & 0.201 & 0.197 & 0.183 & 0.196 & 0.192 & 0.187 & 0.233 & 0.217 & 0.208 \\
Mamba + Selection & 0.113 & 0.188 & 0.186 & 0.176 & 0.183 & 0.181 & 0.173 & 0.232 & 0.211 & 0.207 \\
Mamba + Layer & 0.116 & 0.193 & 0.192 & 0.178 & 0.188 & 0.183 & 0.181 & 0.236 & 0.213 & 0.212 \\
Mamba + Adapter & 0.107 & 0.181 & 0.173 & 0.166 & 0.177 & 0.172 & 0.163 & 0.223 & 0.208 & 0.203 \\
\bottomrule
\bottomrule
\end{tabular}
\end{adjustbox}
\caption{Root Mean Squared Error (RMSE) of predicted social media engagement metrics over a 7-day horizon under counterfactual scenarios for the bushfire dataset.}
\label{tab:rmse_bushfire}
\end{table}

\begin{table}[t]
\centering
\begin{adjustbox}{max width=\linewidth}
\begin{tabular}{l|ccc|ccc|ccc}
\toprule
\toprule
\textbf{Model} & \multicolumn{3}{c|}{\textbf{Scenario 1: Exposure}} & \multicolumn{3}{c|}{\textbf{Scenario 2: Timing}} & \multicolumn{3}{c}{\textbf{Scenario 3: Treatment}} \\
 & 20\% & 40\% & 60\% & -5 days & -3 days & -1 day & 1-day & 3-day & 5-day \\
\midrule
Transformer + Token & 0.522 & 0.447 & 0.383 & 0.462 & 0.413 & 0.392 & 0.518 & 0.461 & 0.443 \\
Transformer + Attention & 0.488 & 0.403 & 0.362 & 0.428 & 0.377 & 0.333 & 0.512 & 0.447 & 0.426 \\
Transformer + Layer & 0.497 & 0.417 & 0.363 & 0.442 & 0.383 & 0.357 & 0.492 & 0.408 & 0.397 \\
Transformer + Adapter & 0.473 & 0.387 & 0.338 & 0.412 & 0.353 & 0.327 & 0.523 & 0.467 & 0.443 \\
Mamba + Token & 0.512 & 0.433 & 0.377 & 0.452 & 0.402 & 0.378 & 0.517 & 0.463 & 0.447 \\
Mamba + Selection & 0.478 & 0.397 & 0.352 & 0.413 & 0.362 & 0.328 & 0.513 & 0.452 & 0.433 \\
Mamba + Layer & 0.483 & 0.407 & 0.353 & 0.427 & 0.372 & 0.342 & 0.517 & 0.453 & 0.432 \\
Mamba + Adapter & 0.467 & 0.378 & 0.337 & 0.398 & 0.352 & 0.318 & 0.507 & 0.447 & 0.418 \\
\bottomrule
\bottomrule
\end{tabular}
\end{adjustbox}
\caption{Binary Cross Entropy (BCE) of treatment intensity predictions over a 7-day horizon for the bushfire dataset across counterfactual scenarios.}
\label{tab:bce_bushfire}
\end{table}

\end{document}